\documentclass[12pt]{report}

\usepackage[a4paper,top=1in,bottom=1in,left=1in,right=1in]{geometry}
\usepackage{amsmath, amssymb, bm}
\usepackage{caption}
\usepackage{graphicx}
\usepackage[colorinlistoftodos]{todonotes}
\usepackage[colorlinks=true, allcolors=black]{hyperref}
\usepackage{float}
\usepackage{setspace}
\usepackage{subfigure}
\usepackage{url}
\usepackage{tabularx}
\usepackage[utf8]{inputenc}
\usepackage{mathptmx} 
\usepackage{titlesec}
\titlespacing{\chapter}{0pt}{0pt}{0pt} 
\usepackage{fancyhdr}
\usepackage{etoolbox}
\usepackage{appendix}
\usepackage{siunitx}
\usepackage{nomencl}

\makenomenclature
\usepackage{caption}
\usepackage{lipsum}
\usepackage{algpseudocode}
\usepackage{empheq}
\usepackage{mdwlist}
\usepackage{booktabs}
\usepackage{colortbl}
\usepackage{moresize}

\usepackage{everypage}
\usepackage{afterpage}
\usepackage{pdfpages}

\def\PageTopMargin{0in}
\def\PageLeftMargin{0in}

\newcommand\atxy[3]{
    \AddThispageHook{
        \smash{
            \hspace*{
                \dimexpr-\PageLeftMargin-\hoffset+#1\relax
            }
            \raisebox{
                \dimexpr\PageTopMargin+\voffset-#2\relax
            }{#3}
        }
    }
}

\newcommand{\insertFig}[3]{
  \begin{figure}[!htb]
    \centering
    \includegraphics[width=#1\textwidth]{#2}
    \caption{#3}
    \label{fig:#2}
  \end{figure}
}

\usepackage[style=ieee]{biblatex}
\DeclareSourcemap{
  \maps[datatype=bibtex]{
    \map{
       \step[fieldsource=title, match=\regexp{\b([A-Z]{2,})\b}, replace={{}{$1}}]
    }
  }
}
 
\DeclareCaptionType{appfigure}[Figure]
\DeclareCaptionType{apptable}[Table]

\newcommand{\chapname}{Chapter }
 
\titleformat{\chapter}[block]
{\bfseries\LARGE\centering}
{}{1em}{}[\rule{\textwidth}{0.3pt}]

\titleformat{\section}
{\bfseries\large}
{\thesection}{1em}{}

\titleformat{\subsection}
{\normalfont\bfseries}
{\thesubsection}{1em}{}

\titleformat{\subsubsection}
{\normalfont\bfseries}
{\thesubsubsection}{1em}{}

\usepackage{fancyhdr}
\usepackage{multirow}
\usepackage{multicol}

\usepackage{placeins}

\makeatletter
\renewcommand*\env@matrix[1][*\c@MaxMatrixCols c]{
    \hskip -\arraycolsep
    \let\@ifnextchar\new@ifnextchar
    \array{#1}}
\makeatother

\usepackage{enumitem}

\usepackage{listings}
\usepackage{color}
 
\definecolor{codegreen}{rgb}{0,0.6,0}
\definecolor{codegray}{rgb}{0.5,0.5,0.5}
\definecolor{codepurple}{rgb}{0.58,0,0.82}
\definecolor{backcolour}{rgb}{0.95,0.95,0.95}

\newcommand{\codesize}{\fontsize{10pt}{11pt}\selectfont}

\lstdefinestyle{mystyle}{
    backgroundcolor=\color{backcolour},   
    commentstyle=\color{codegreen},
    keywordstyle=\color{magenta},
    numberstyle=\tiny\color{codegray},
    stringstyle=\color{codepurple},
    basicstyle=\ttfamily\codesize,
    breakatwhitespace=true,         
    breaklines=true,                 
    captionpos=b,                    
    keepspaces=false,                 
    numbers=left,                    
    numbersep=5pt,                  
    showspaces=false,                
    showstringspaces=false,
    showtabs=false,                  
    tabsize=2,
    showlines = true,
    fontadjust = true,
    framexleftmargin = 10 pt,
    resetmargins = true,
    basewidth = 0.5em
}
 

\usepackage{multicol}
 
\newcommand{\C}{\mathbb{C}}

\usepackage{makecell}
\usepackage{emptypage}

\newcolumntype{R}{>{\raggedleft\arraybackslash}X}
\newcolumntype{L}{>{\raggedright\arraybackslash}X}
\newcolumntype{C}{>{\centering\arraybackslash}X}

\usepackage{textcomp}

\begin{document}
\sloppy

\pagenumbering{gobble}
\newgeometry{top=1.4in,bottom=1in,left=2.5in,right=1.5in} 

\atxy{0in}{0in}{
    \rotatebox[origin=l]{-90}{
        \makebox[9.75in]{
            \parbox{9.75in}{
                \hrulefill\\
                \rotatebox[origin=l]{90}{
                    \parbox{0.9in}{
                        \vspace{2.5em}
                        Acad. Year\\2023-2024\\[1ex]
                        Project No.\\C141
                    }
                }\\[-0.7in]
                \centerline{\textbf{\large%
                \hspace{5in} 
                \parbox{9.75in}{%
                Behavior Imitation for Manipulator Control and Grasping \\
                with Deep Reinforcement Learning
                }}}
                \rule[0pt]{0pt}{0pt}
            }
        }
    }
}

\begin{center}
\linespread{1}\huge\textbf{Behavior Imitation for Manipulator Control and Grasping with Deep Reinforcement Learning}
\vspace{1.2in}
\end{center}

\begin{figure}[h]
    \centering
    \includegraphics[width=0.8\textwidth]{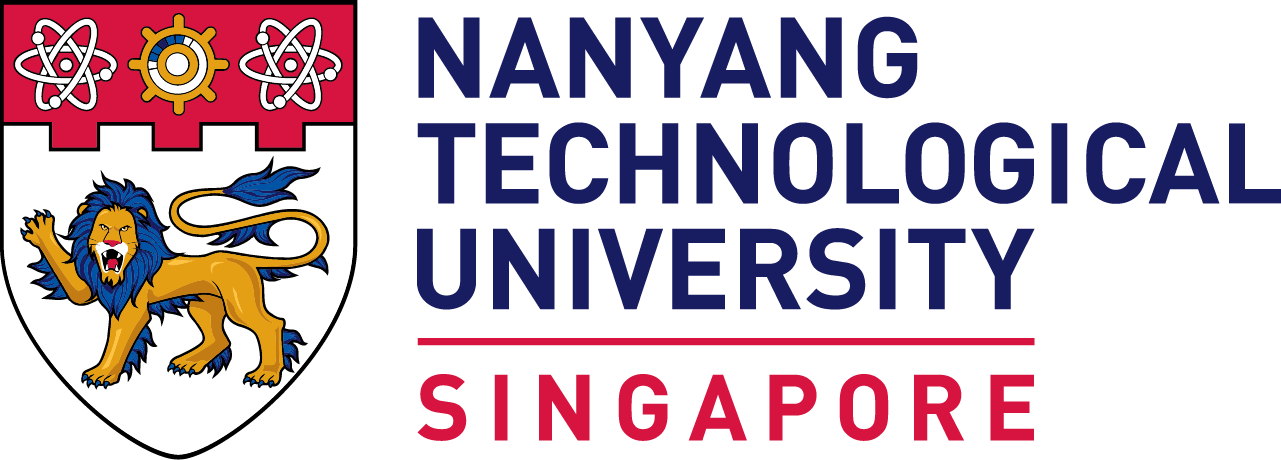}
\end{figure}

\begin{center}
\vspace{1.2in}
\textbf{Liu Qiyuan}
\vspace{1.2in}

\textbf{School of Mechanical \& Aerospace Engineering\\Nanyang Technological University, Singapore}
\vspace{1.2in}

\textbf{Year 2023/24}
\end{center}
\clearpage
\restoregeometry

\begin{titlepage}
    \begin{center}
    
        BEHAVIOR IMITATION FOR MANIPULATOR CONTROL AND GRASPING \\ WITH DEEP REINFORCEMENT LEARNING\\[1.8in]
    
    SUBMITTED BY\\
    LIU QIYUAN\\
    (U2020275D)\\[1.8in]
    
    SCHOOL OF MECHANICAL \& AEROSPACE ENGINEERING\\[1.8in]
    
    A final-year project report presented to\\
    Nanyang Technological University, Singapore\\
    in partial fulfilment of the requirements for the degree of\\
    Bachelor of Engineering (Mechanical Engineering)\\
    Nanyang Technological University, Singapore\\[0.3in]
    
    Year 2023/24
    \end{center}
    \newpage
    \end{titlepage}
\pagenumbering{roman}
\pagestyle{fancy}
\fancyhf{}
\cfoot{\thepage}


\chapter*{Abstract}
\rhead{Abstract}
\addcontentsline{toc}{chapter}{Abstract}


\vspace{1cm}
\noindent The existing Motion Imitation models typically require expert data obtained through MoCap devices, but the vast amount of training data needed is difficult to acquire, necessitating substantial investments of financial resources, manpower, and time. This project combines 3D human pose estimation with reinforcement learning, proposing a novel model that simplifies Motion Imitation into a prediction problem of joint angle values in reinforcement learning. This significantly reduces the reliance on vast amounts of training data, enabling the agent to learn an imitation policy from just a few seconds of video and exhibit strong generalization capabilities. It can quickly apply the learned policy to imitate human arm motions in unfamiliar videos. The model first extracts skeletal motions of human arms from a given video using 3D human pose estimation. These extracted arm motions are then morphologically retargeted onto a robotic manipulator. Subsequently, the retargeted motions are used to generate reference motions. Finally, these reference motions are used to formulate a reinforcement learning problem, enabling the agent to learn policy for imitating human arm motions. This project excels at imitation tasks and demonstrates robust transferability, accurately imitating human arm motions from other unfamiliar videos. This project provides a lightweight, convenient, efficient, and accurate Motion Imitation model. While simplifying the complex process of Motion Imitation, it achieves notably outstanding performance.


\par
\vspace{1cm}
\textbf{Keywords: Motion Imitation, Imitation Learning, Deep Reinforcement Learning, 3D Human Pose Estimation, Motion Retargeting, Inverse Kenimatics, PyBullet Simulation.} 

    
\newpage


\chapter*{Graphical Abstract}
\rhead{Graphical Abstract}
\addcontentsline{toc}{chapter}{Graphical Abstract}
        
\insertFig{1.0}{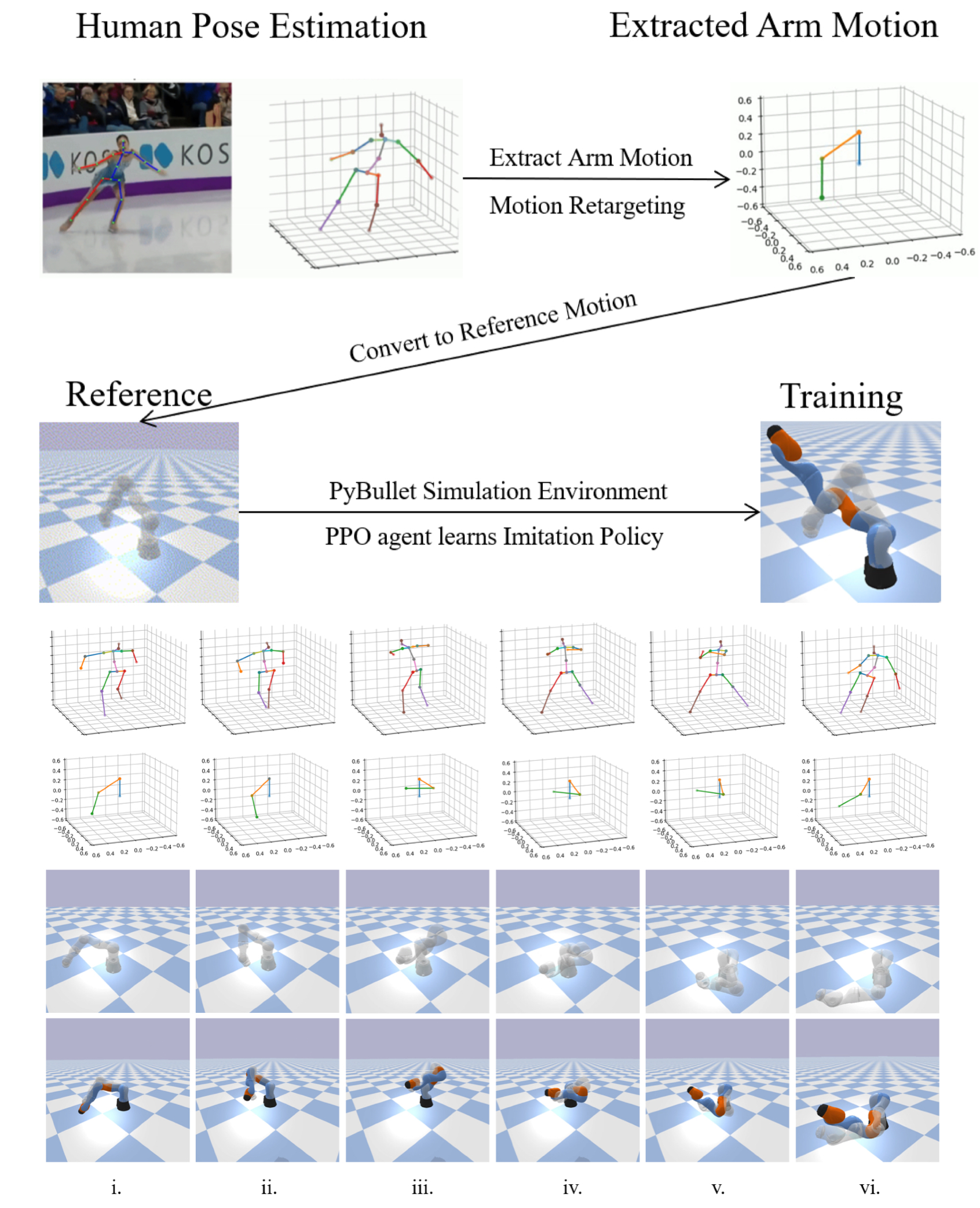}{Graphical Abstract}

    
\newpage


\chapter*{Acknowledgement}
\rhead{Acknowledgement}
\addcontentsline{toc}{chapter}{Acknowledgement}

\noindent I would like to express my heartfelt gratitude to Nanyang Technological University, Singapore (NTU) for providing me with a platform rich in resources and opportunities, enabling me to contribute to the forefront of technology.\\

\noindent I also would like to extend my sincere thanks to my Supervisors, Prof. Wen Bihan and Prof. Lyu Chen, for their timely and effective guidance, setting a fine example of scientific research for me to follow.\\

\noindent In addition, I am grateful to the friends I have met at university who share my aspirations. Thank you all for your dedication and companionship, which have made my university life a wonderful journey.\\

\noindent I extend my deepest appreciation to my parents back in China. As my unwavering support system, they have been there for me every step of the way, providing me with constant motivation over the past four years.\\

\noindent Lastly, special thanks to Ms. Zhan Jie. I would like to quote the only line from Shakespeare's sonnets that I know: “Love alters not with his brief and weeks. But bears it out even to the edge of doom."\\

\noindent Everyone I've mentioned has helped me improve, grow, and become a better person. Thank you all for filling me with pride and inspiring me to continue pursuing excellence in all I do.\\

\newpage
\setcounter{tocdepth}{2}

\tableofcontents
\rhead{Table of Contents}
\newpage

\renewcommand{\listfigurename}{Lists of Figures}
\rhead{Lists of Figures}
\listoffigures 
\addcontentsline{toc}{chapter}{Lists of Figures}

\newpage

\listoftables 
\addcontentsline{toc}{chapter}{Lists of Tables}
\rhead{Lists of Tables}
\newpage

\rhead{}

\pagenumbering{arabic}

\lhead{Introduction}
\chapter{Introduction}

\section{Background and Motivation}

\noindent It is challenging for existing Motion Imitation models to obtain expert data, as there is a need to collect a significant amount of expert demonstration data, which can be costly and time-consuming \parencite{8917306}. To be more specific, the requirement for a large volume of expert data is particularly challenging in high-dimensional environments with complex dynamics \parencite{garg2022iqlearn}. For example, in learning agile robotic locomotion skills by imitating animals, the expert data is obtained using Motion Capture (MoCap) devices on animals \parencite{RoboImitationPeng20}. Although MoCap devices can offer high-quality expert data, the MoCap process itself can be expensive, in terms of both its initiating cost and the time invested in recording animal motion. How to simplify the Motion Imitation process so that it can efficiently imitate motions using a small amount of easily accessible data has become a research area worth exploring.\\

\noindent However, the integration of Deep Reinforcement Learning (DRL) for robotic manipulator control policy learning with 3D Human Pose Estimation (3D HPE), provides a novel and interdisciplinary approach for Motion Imitation, which simplifies the process of obtaining expert data by utilizing recorded human arm videos.\\

\noindent \textbf{Human Pose Estimation.} Recent rapid developments in Computer Vision (CV) have led to significant progress in 3D Human Pose Estimation (3D HPE) from monocular images and videos. Techniques such as DeepMimic that combines data-driven behavior and physical simulation using deep neural network for motion reconstruction \parencite{2018-TOG-deepMimic}, graph and temporal convolutional networks for multi-person pose estimation \Parencite{cheng2021graph}, and strided transformer network that exploiting temporal contexts to estimate 3D human pose from 2D key points motion \parencite{li2023exploiting}, have showcased the capability to accurately estimate 3D human poses. These techniques provide a foundation for extracting accurate human arm motion data, which can serve as expert's demonstration data to guide the robotic manipulator's motion imitation process.\\

\noindent \textbf{DRL for Robotics Control.} Moreover, the field of Deep Reinforcement Learning (DRL) has seen astonishing growth, especially in the context of robotics intelligence control. DRL models such as Deep Q-Network (DQN) \parencite{mnih2013playing}, Twin Delayed Deep Deterministic Policy Gradient (TD3) \parencite{fujimoto2018addressing}, and Proximal Policy Optimization (PPO) \parencite{schulman2017proximal}, have demonstrated efficiency and accuracy in learning complex control policies and decision-making tasks. Additionally, the application of deep reinforcement learning in diverse domains underlines the potential for controlling robotic manipulators. \parencite{10.3390/s21041278}\\

\section{Objective}

\noindent The objective of this project is to develop a novel Motion Imitation model that builds upon the progress in 3D Human Pose Estimation and Deep Reinforcement Learning, enabling robotic manipulators to effectively imitate human arm motion using One-Shot from any valid input videos. To be more specific, the results from 3D human pose estimation will be leveraged to guide a deep reinforcement learning agent in obtaining a control policy that allows the robotic manipulator to imitate the demonstrated human arm motion. Eventually, the results of motion imitation will be showcased in a simulation environment. Furthermore, this project may benefit the development of imitation learning, robotic manipulation, and human-robot interaction, shedding light on the broader implications and applications of such innovative work.\\

\section{Scope}

\noindent The scope of this project involves applying a Strided Transformer Network \parencite{li2023exploiting} to reconstruct 3D human skeleton motion estimation based on the demonstrated video. The raw motion results obtained are subsequently processed and translated into expert data using Motion Retargeting \& Inverse Kinematics. This expert data is then input into a Proximal Policy Optimization \parencite{zhang2020proximal} agent with a reward function specifically designed for Motion Imitation \parencite{RoboImitationPeng20}, guiding the learning process to acquire a control policy that enables the robotic manipulator to effectively imitate human arm motion from input video. Finally, a PyBullet simulation environment is constructed to demonstrate the results of imitation.\\

\newpage

\lhead{Literature Review}
\chapter{Literature Review}

\section{3D Pose Estimation}

\noindent 3D pose estimation refers to the process of estimating the 3D position and orientation of an object, animal, or human body from the given images or videos. It is a complex problem in the field of Computer Vision, involving the estimation of the spatial, or potentially, temporal configuration of the target. Various approaches have been proposed to address this challenge, including DeepPose utilizing Convolutional Neural Network (CNN) to learn human pose representations directly from images \parencite{Toshev_2014}, OpenPose integrating a multi-stage Convolutional Neural Network with geometrical transformation to detect and localize key points of human body in images or videos \parencite{cao2019openpose}, and Strided Transformer exploiting temporal context of human motion in a strided manner \parencite{li2023exploiting}. In the context of Motion Imitation, compared with MoCap devices, 3D pose estimation provides a rather convenient and flexible method to obtain expert motion data. Although MoCap devices offer high-accuracy motion data, they can be expensive and require a controlled environment. In contrast, 3D human pose estimation is more flexible, especially for the complex imitation task where a large volume of expert data is required.\\

\noindent \textbf{DeepPose. \& OpenPose.} Both DeepPose and OpenPose are CNN-based approaches, which significantly benefit the development in the field of 3D pose estimation. DeepPose architecture consists of multiple layers of convolutional and recurrent neural networks, followed by fully connected layers. The network parameter is trained on large datasets of labeled human motion data, where each label includes the key point 3D coordinate information. Depending on the specific application, DeepPose may leverage additional information such as temporal dependencies in video sequences or context from surrounding frames to improve estimation accuracy \parencite{Toshev_2014}. Similar to DeepPose, OpenPose is an updated version with multi-state CNN suitable for the estimation of multi-person motion. Moreover, OpenPose incorporates depth information from stereo images or videos, further improving the accuracy of estimated motion \parencite{cao2019openpose}.\\

\noindent \textbf{Strided Transformer.} Strided Transformer Network is originally applied in Natural Language Processing tasks. However, recent advancements have introduced transformer-based models that can effectively exploit spatial and temporal context to lift 2D estimated motion to 3D \parencite{qiu2022ivt}. Integrating with transformer layers, the model can extract both local and global spatial and temporal relationships between key points, which result in improved estimation accuracy \parencite{li2023exploiting}. Compared with DeepPose \& OpenPose, the transformer architecture allows for extracting global information, which infers the model can potentially understand the relationship between the motion of body parts and whole, resulting in more accurate estimation \parencite{10.48550/arxiv.2205.15448}. Moreover, depending on the specific implementation, the transformer can be more computationally efficient than conventional CNN-based models \parencite{10.1007/s11263-019-01250-9}.\\

\insertFig{0.9}{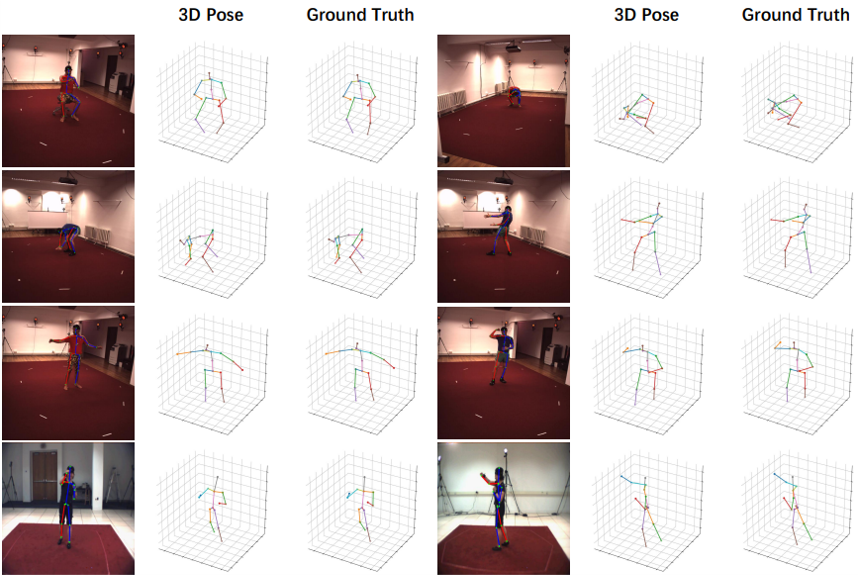}{Strided Transformer Network for Human Pose Estimation \parencite{li2023exploiting}}

\noindent For example, as shown in \hyperref[fig:Fig/hpe.png]{Figure 2-1} Li et al. utilized a strided transformer network trained a motion estimation model to extract 3D human motion \parencite{li2023exploiting}, emphasizing the possibility of using the extracted motion as expert data to train a motion imitating robotic manipulator control policy. To be more specific, \hyperref[fig:Fig/hpe.png]{Figure 2-1} showcases a Strided Transformer model, where the output motion after minor modification is applicable as reference motion to robotic manipulator motion imitation tasks.\\

\section{Motion Retargeting}

\noindent Motion Retargeting refers to a technique commonly applied in animation or robotics to transfer the same motion between characters. The fundamental idea behind motion retargeting is to utilize motion data obtained from one source, such as MoCap or 3D Pose Estimation, and adapt it to fit different morphology. This adaptation involves adjusting the motion to match proportions, skeletal structure, and range of motion while maintaining the original kinematics, dynamics, and style of motion \parencite{Aberman_2019}. In the context of the Motion Imitation task, the motion retargeting process adapts human arm morphology to fit that of robotic manipulators.\\

\noindent Motion retargeting can be achieved through various methods, including traditional geometrical transformation, inverse kinematics \parencite{10.1145/280814.280820}, and modern data-driven deep learning techniques \parencite{10.1145/3386569.3392462}. Traditional methods address motion retargeting problems by solving a series of inverse forward kinematic equations that describe the relationships between connected segments of a skeletal structure. The solved relationships are transferred to another skeletal structure with geometrical transform and forward kinematics. In contrast, modern techniques involve recurrent neural networks and generative adversarial networks, which are data-driven approaches that require a large volume of labeled training data for training to achieve desired accuracy.\\

\insertFig{0.9}{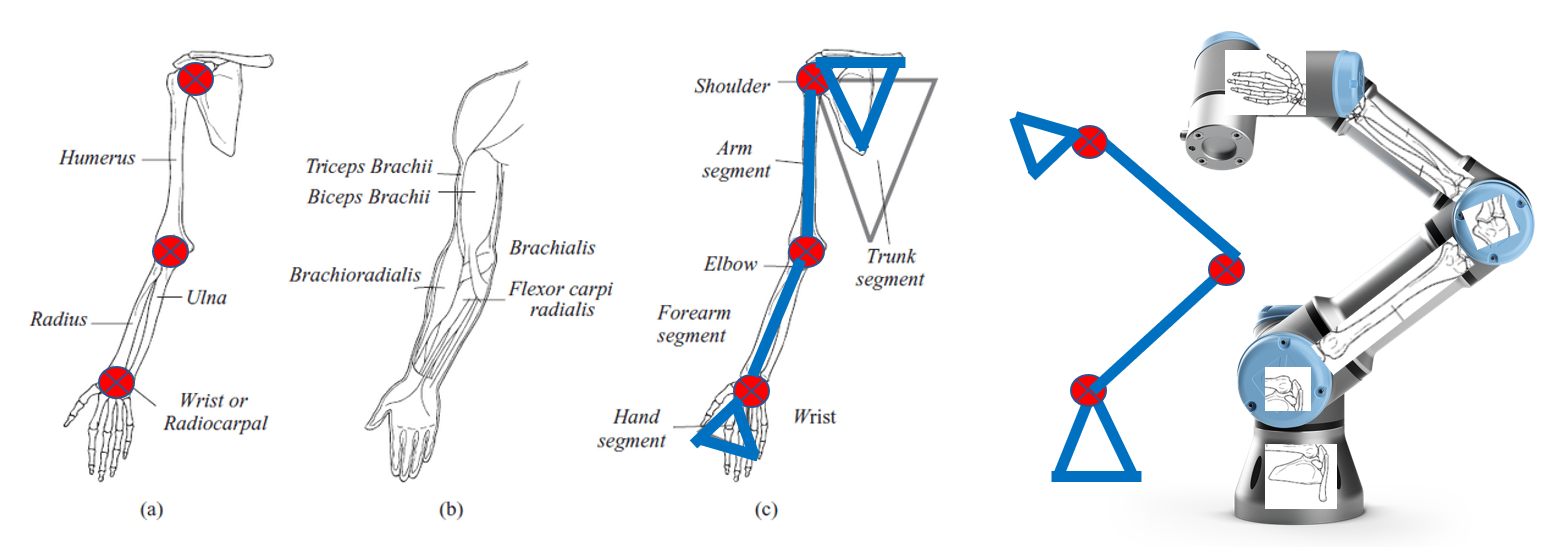}{Retargeting human arm morphology to robotic manipulators}

\noindent In the context of Motion Imitation, the knowledge, i.e., dynamics, skeletal structure, and feasible workspace, of robotic manipulators is known. Therefore, as shown in \hyperref[fig:Fig/retargeting.png]{Figure 2-2}, with the known knowledge, traditional methods are more efficient for retargeting expert arm motion to that of robotic manipulators.\\

\section{Reinforcement Learning}

\noindent Reinforcement Learning (RL) is an interactive intelligent agent, as shown in \hyperref[fig:Fig/drl_illustration.png]{Figure 2-3} where it learns a policy to make decisions by interacting with an environment and receiving feedback in terms of observation and reward \parencite{Kaelbling.1996}. Integrated with deep neural networks (DNN), Deep Reinforcement Learning (DRL) enables agents to learn a more sophisticated control policy in a more dynamic environment, achieving human-level control in various domains \parencite{Mnih.2015}. For the tasks of Motion Imitation, a DRL problem can be formulated to train a control policy that imitates target expert motion.\\

\insertFig{0.9}{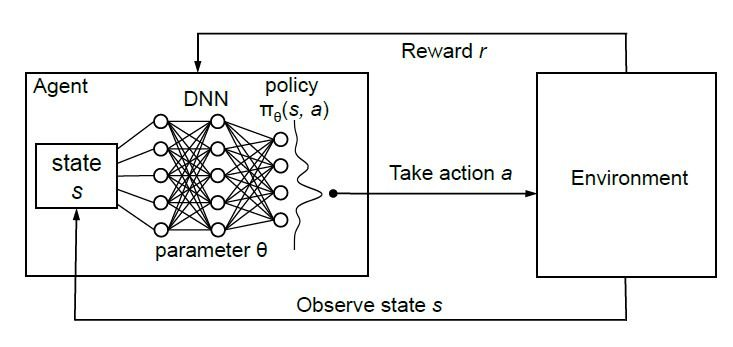}{Deep Reinforcement Learning Schematic Process}

\noindent \textbf{DQN.} Deep Q-Network combines DNN and Q-learning \parencite{mnih2013playing} and has been widely applied in various domains. Although DQN has successfully solved complex problems, its non-linearity and discrete nature led to the challenge of control tasks in continuous action space.\\

\noindent \textbf{TD3.} Twin Delayed Deep Deterministic Policy Gradient \parencite{fujimoto2018addressing} successfully resolved DQN's non-linearity and discrete nature. It has been widely applied in various fields owing to its success in addressing continuous control tasks, making it suitable for control tasks in various domains, particularly for robotic manipulator control. However, TD3 performance can be sensitive to hyperparameters, requiring a crucial hyperparameter tuning process for achieving the desired performance. Moreover, similar to other Q-Learning-based models, TD3 suffers from overestimation bias, which can lead to sub-optimal policies.

\noindent \textbf{PPO.} Proximal Policy Optimization has been increasingly applied to control robotic manipulators PPO has been benchmarked on typical tasks and shown superior performance.\parencite{schulman2017proximal}. Compared to TD3, PPO directly optimizes the policy through a clipped objective function, without explicitly estimating the value function, which avoids the overestimation issue of TD3. Moreover, PPO is more stable in training without the requirement to carefully tune hyperparameters, especially in complex environments.\\

\noindent Based on the relevant literature of the aforementioned trending DRL algorithms, PPO has demonstrated superior adaptability in complex environments, making it well-suited for addressing the challenges posed by complex and dynamic environments in imitation of the human expert pose. Furthermore, PPO has been shown to perform better in solving continuous action space control tasks of robotic manipulators, making it an even more favorable choice for imitating human pose.\\

\section{Motion Imitation}

\noindent Imitation learning refers to the process in which an agent learns a control policy by imitating expert demonstrations. Imitation learning integrated with DRL (Deep Imitation Learning) shows a promising future for autonomous robot manipulation. It does not rely on hard-coded control policy, instead, allowing robots to learn complex manipulation tasks from the expert demonstration. The emergence of various imitation learning approaches in controlling robotic manipulators, compared to hard-coded ones, granted it enough flexibility to explore potential trajectories while fulfilling the task of imitating specified expert demonstration \parencite{Jung.2021}. This showcased its potential to address the limitations of traditional intelligence control approaches and enable robotic manipulators to achieve more complex manipulations.\\

\noindent \textbf{BC.} Behavior Clone is an imitation learning approach where robots learn policies without any trials by imitating expert demonstration, but it learns a direct mapping from the expert demonstration to the action performed by the robot \parencite{torabi2018behavioral}. However, learning end-to-end direct mapping means it is highly sensitive to errors in expert demonstrations, and it also means BC will suffer from compounding errors and lack to explore and learn from its interactions with the environment, hindering the robot's capability to discover potentially superior policies.

\noindent \textbf{IRL.} Inverse Reinforcement Learning is a technique for learning and comprehending the underlying reward function from expert demonstrations, enabling the agent to learn from the behavior of expert demonstrations instead of explicit reward signals \parencite{Abbeel.2004}. IRL has been applied to various domains to transfer task knowledge from an expert to a robot in a dynamic environment and acquire the ability to imitate expert demonstration \parencite{Hansen.2020}. However, IRL may suffer from computational drawbacks, particularly when applied in complex tasks, as it requires solving a Markov Decision Problem to obtain the reward function, which can be computationally expensive. Therefore, it may fail to learn a reasonable behavior in certain complex environments.\\

\noindent \textbf{GAIL.} Generative Adversarial Imitation Learning combines generative adversarial networks with deep reinforcement learning to learn to generate policy that closely imitates expert demonstration \parencite{ho2016generative}. It is applied to various domains and has shown its outstanding capability of learning human-like behaviors in robotic manipulator control. However, GAIL may fail to generate a reasonable imitation policy when the available demonstrations are limited or not representative. Therefore, it requires a large volume of expert demonstrations covering enough sample cases.\\

\noindent \textbf{Motion Imitation.}\\

\insertFig{0.9}{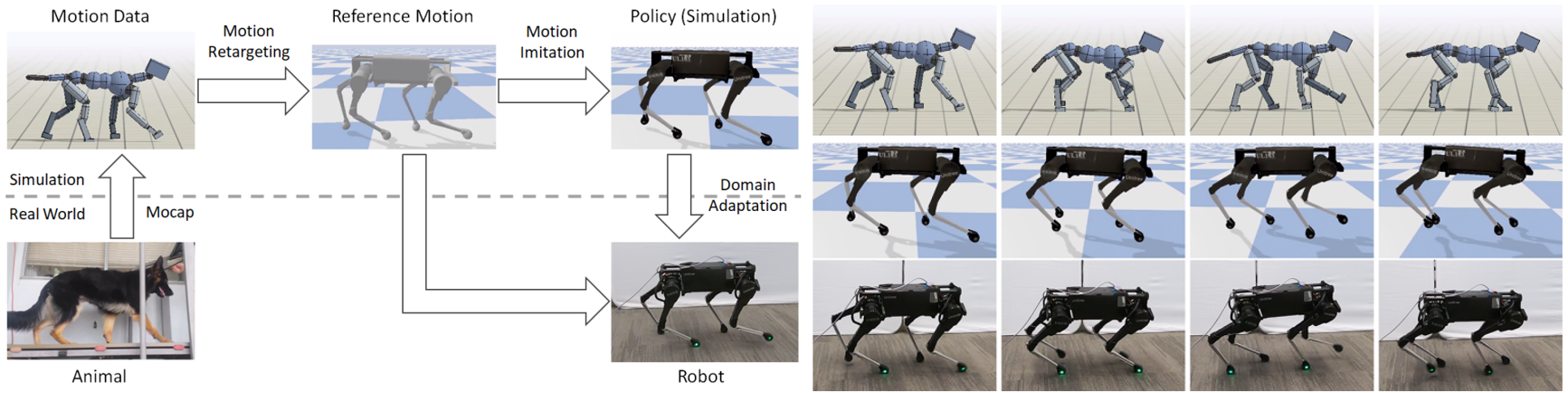}{Framework and result of Motion Imitation \parencite{RoboImitationPeng20}}

\noindent Besides the aforementioned trending imitation learning approaches, direct Motion Imitation by formulating a regular DRL problem (schematic illustration shown in \hyperref[fig:Fig/motion_imitate.png]{Figure 2-5}) can be an effective technique to train the intelligent agent, obtaining a control policy that can imitate expert demonstration data \parencite{RoboImitationPeng20}. This method can be decomposed into the following steps.\\

\noindent \textbf{Key-Point Retargeting.} To effectively obtain expert data, Motion Imitation utilized a MoCap device on an animal to record the key points' trajectories. These key points on animals should be defined such that they can represent specific anatomical landmarks and motion characteristics \parencite{Zhao.2022}. In motion imitating, the key points are defined as respective joints to capture the complete motion of the animal expert.\\

\noindent \textbf{Motion Retargeting.} When using the motion data captured from animals, there is a common problem that the subject's morphology in terms of proportions and skeletal structure tends to differ from that of the robot. In such cases, Inverse-Kinematics is applied to mitigate this discrepancy \Parencite{10.1145/280814.280820}.\\

\noindent \textbf{Motion Imitation.} The task of imitation can be formulated as a typical DRL problem. The objective is to learn a control policy that enables the agent to minimize the discrepancy between the robotic manipulator’s motion and reference motion. This discrepancy is reflected by a specifically designed reward function that encourages the policy to imitate target reference motion. Training through trials and error, the agent eventually converges to an optimal policy that completes the Motion Imitation.\\

\section{PyBullet Simulation Environment}

\noindent The simulation environments refer to a virtual space where simulations are conducted to reflect real-world system behaviors, without the need to directly interact with or affect the real system. Moreover, a simulation environment offers a controlled and cost-effective approach to explore and understand the behavior of a complex system without the risks or expenses associated with real-world experimentation.\\

\noindent PyBullet is an open-source physical simulator widely used for simulating robotic systems, particularly in the context of Deep Reinforcement Learning. It provides accurate physics simulation, allowing robotic manipulators to interact with the environment realistically which benefits model deployment. Moreover, PyBullet is designed for efficiency, allowing for fast simulation of complex robotic manipulator control systems. Most importantly, PyBullet is compatible with various operating systems, reducing the complexity for reproduction \Parencite{coumans2019pybullet}.

\newpage

\lhead{Methodology}
\chapter{Methodology}

\section{Motion Imitation Overview}

\insertFig{0.75}{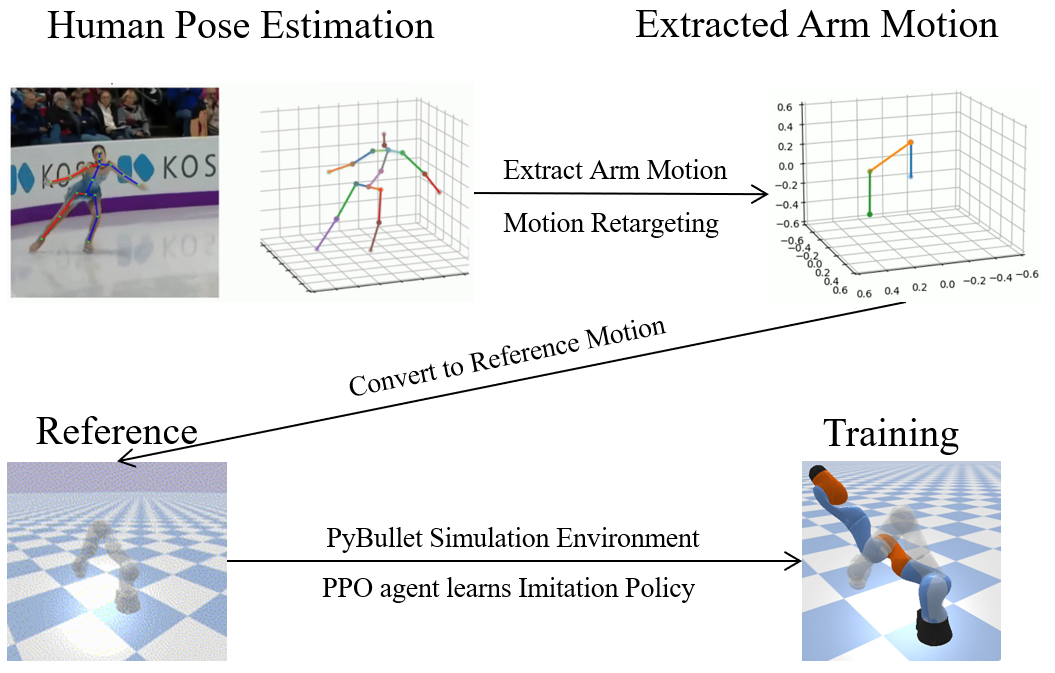}{Overview of Motion Imitation project flow} 

\noindent The objective of this project is to provide a model that enables robotic manipulators to imitate human arm motion from any human full-body motion videos. As shown by the overview of Motion Imitation project flow in \hyperref[fig:Fig/overview.png]{Figure 3-1}, the process consists of three main stages: raw motion extraction, motion retargeting, and motion imitating. In the first stage, the model receives as input a full-body human motion video and extracts the 3D raw arm motion from it, with a combined network of yolov3 \Parencite{redmon2018yolov3}, HRNet \Parencite{wang2020deep}, and strided transformer network \Parencite{li2023exploiting}. Subsequently, in the middle stage, the extracted raw arm motion is retargeted to match the robotic manipulator’s morphology with geometric transformation and inverse kinematics, providing comprehensive reference motion to the robotic manipulator. Finally, in the last stage, the retargeted reference motion is used to formulate a DRL motion imitation problem, where the control policy is trained to imitate reference motion. To enhance the robustness and the generalization ability of the model, the control policy does not generate a series of trajectory. Instead, it focuses on reproducing joint position and end effector position at current timestep.\\

\section{Extract Raw Arm Motion}

\noindent The raw arm motion extraction process adopts the model proposed by Li et. al. \Parencite{li2023exploiting}. The overall process is organized into two main steps: the first step attaches key points that represent human motion frame by frame in a 2D manner and generates the corresponding 2D coordinates of those key points on each frame. Subsequently, in the second step, a strided transformer is introduced to lift the 2D key point coordinates into 3D, and generates the final estimated 3D data of human motion. The two steps of the overall process are described in \hyperref[fig:Fig/hpe_demo.png]{Figure 3-2}.\\

\insertFig{0.75}{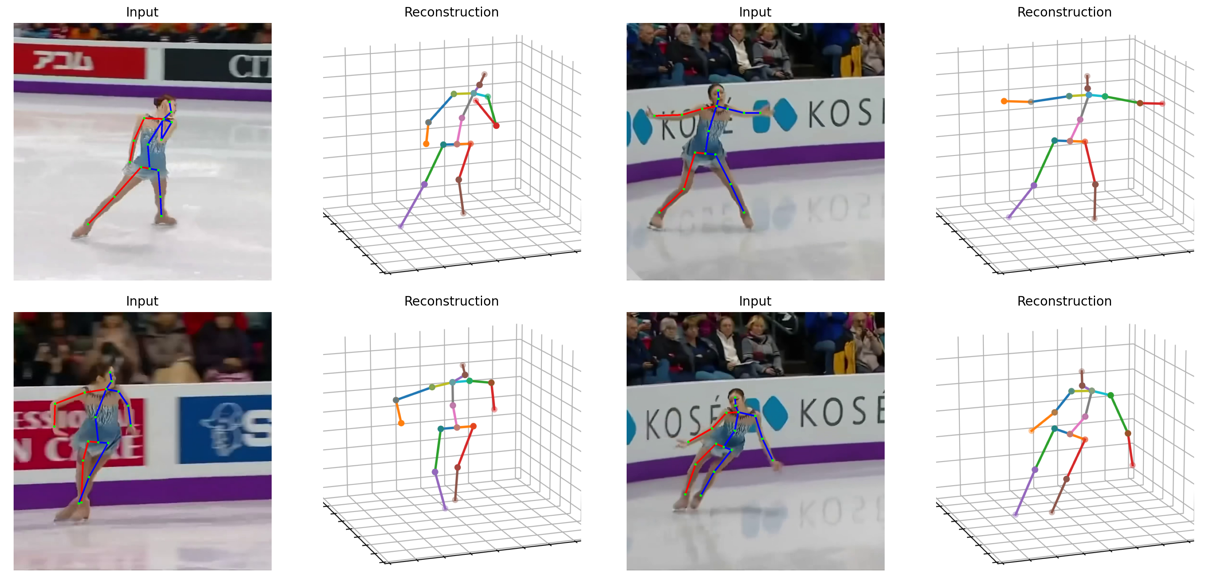}{Raw human motion extraction process} 

\noindent \textbf{Estimate 2D Motion.} This step combined a yolov3 and an HRNet detection model, pretrained to adapt to the requirement of attaching key points to each frame of the input video and generate corresponding coordinates in 2D space. It is worth noting that the selection of locations where key points are attached is crucial. The key points are selected such that they can completely define a human body motion without any ambiguity. Therefore, these key points are defined at the joints, limb extremities, and center of mass of the body, which gives 17 representative key points. In summary, this step takes as input the human motion video and produces as output the motion of 17 key points in terms of coordinates of each point at every frame.\\

\noindent \textbf{Elevate 2D motion to 3D.} This step introduces strided transformer network to lift the 2D motion obtained in the previous step to 3D space. In previous steps, the 17 key points are simply attached to a corresponding target location on every frame, which means the output coordinates are pixel-wise coordinates. However, this lifting step requires reconstructing the entire skeletal structure in a 3D space, where the coordinates are defined in a cartesian coordinate system. Therefore, the transformer network is trained with Human3.6M \Parencite{ionescu2014human3}, a benchmark dataset for human pose estimation, in order to elevate 2D coordinates in pixel-wise context to 3D coordinates in a cartesian coordinates system. The extracted raw motion is grounded with respect to the lowest among all 17 key points to improve the stability of the extracted motion. The reconstruction result is visualized in \hyperref[fig:Fig/hpe_demo.png]{Figure 3-2} on the right, which presents a satisfactory correspondence to the original 2D pixel-wise key points on the left.\\

\noindent \textbf{Extract 3D Arm Motion.} As the training dataset, Human3.6M, contains all 17 key points coordinate information for each labeled data, the resulting reconstructed 3D human motion shares the same skeletal structure that contains all 17 key points. However, for the Motion Imitation task of this project, only 4 interested points, as demonstrated in \hyperref[fig:Fig/retargeting.png]{Figure 2-2}, are needed to represent the arm motion to be imitated. These points are selected as they are highly associated with the base and joints of the robotic manipulator and can fully represent human arm motion.\\

\insertFig{0.8}{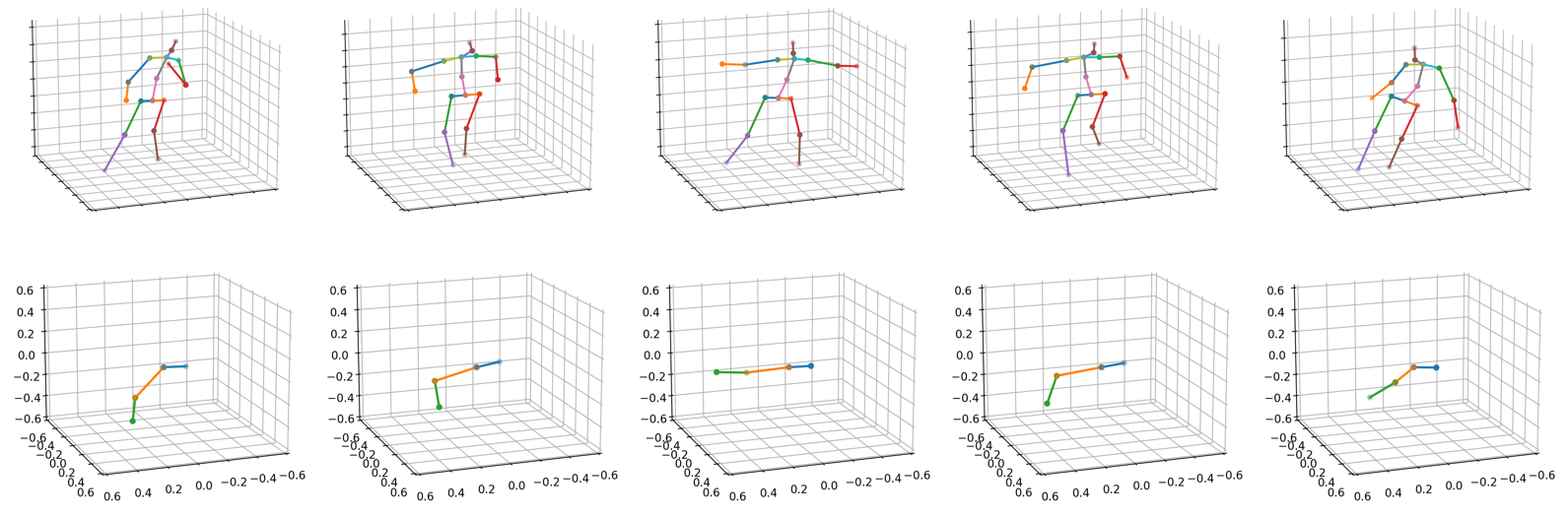}{The extracted arm motion from original skeletal structure} 

\noindent \hyperref[fig:Fig/ext_arm.png]{Figure 3-3} shows the extracted arm motion from the elevated skeletal structure. The extracted arm motion fixes the shoulder joint as the base coordinate, and the succeeding two key points' motion fully characterizes the original arm motion. Although the extraction process transformed the 4 interested key points from a global skeletal cartesian coordinate system to a newly defined local arm cartesian coordinate system, the relative kinematic and its corresponding dynamic share the same relationship but are scaled to a different scale. However, so long as the style of motion, i.e. the relative kinematic and dynamic relationship, is maintained the same, the scaling issue can be easily addressed in the motion retargeting process.\\

\section{Retarget Reference Motion}

\noindent Although the raw motion data that characterizes the human arm motion is extracted from the input expert demonstration video, it is infeasible to directly apply the obtained raw data to Motion Imitation as reference motion. The reason is that the raw arm motion morphology in terms of proportion, orientation, skeletal structure, and kinematics relationship, does not match that of robotic manipulators. Therefore, pre-processing is needed to retarget human arm morphology to match that of the robotic manipulators. The overall method to complete the retargeting process is shown in \hyperref[fig:Fig/mot_ret.png]{Figure 3-4}. To sum up, the motion retargeting process is a preprocessing that takes raw human arm reference motion as input and generates Motion Imitation reference motion.\\

\insertFig{0.9}{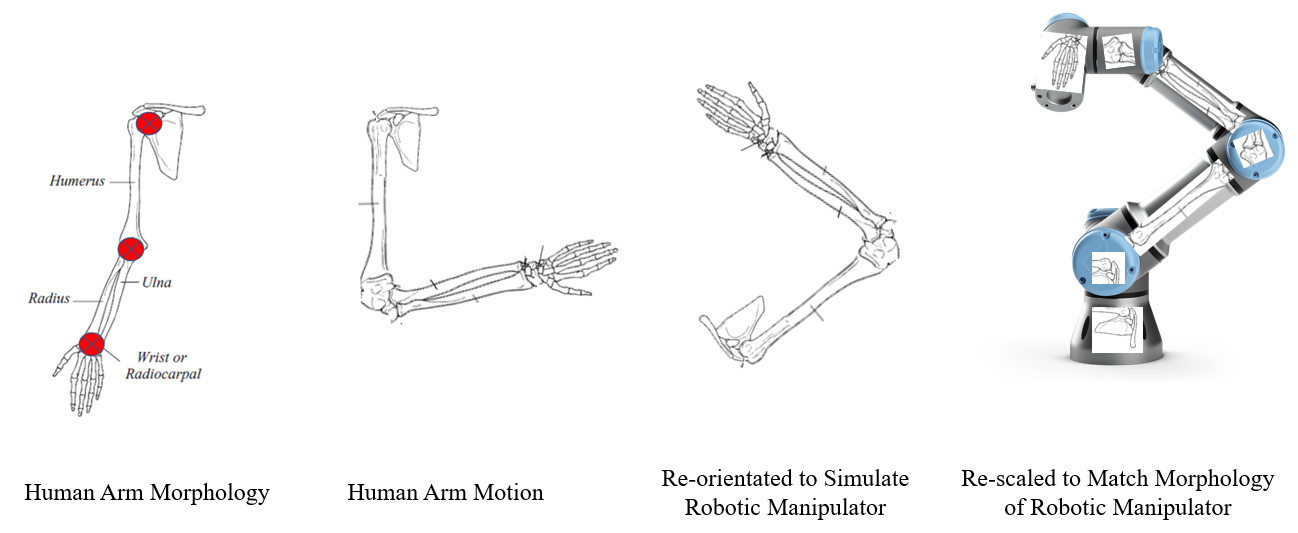}{Align human arm morphology to that of robotic manipulator} 

\noindent Before the motion retargeting process, it is essential to understand the raw reference motion, especially for understanding how raw reference relates to human arm motion in video input and reference motion of robotics manipulators. Because the transformations performed in the motion retargeting process essentially determine a mapping relationship to map a human arm structure to a robotic manipulator. The 4 interested key points are selected in corresponding to the robotic manipulator's joint assembly as shown in \hyperref[fig:Fig/mot_ret.png]{Figure 3-4}, suggesting that the human arm and robotic manipulator share similar skeletal structures. Therefore, the motion retargeting process of this project fully utilizes this significant correspondence, attempting to match the human arm joint and robotic manipulator joint assembly respectively. Once all desired joints and joint assemblies are properly matched, Inverse Kinematics is applied such that it takes all matching pairs of joints as constraints to solve a feasible trajectory that guarantees the exact positional equivalency between the reference motion endpoint and end-effector, but at the same time maximize the similarity between human arm motion and reference motion. It is worth noting that, inverse kinematics is introduced to solve a feasible trajectory with the robotic manipulator’s workspace, in case some human arm motions are too complex for the robotic manipulator to imitate, where the complex motion typically refers to those motions outside the robotic manipulator workspace even after motion retargeting process as the range of human joint is much larger than that of robotic manipulators.\\

\insertFig{0.9}{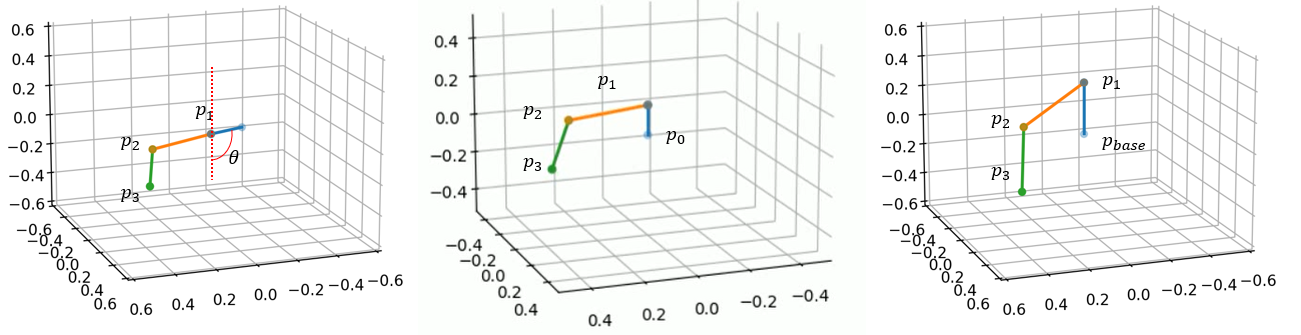}{Base matching \& link rescaling process} 

\noindent \textbf{Base Matching.} The extracted raw reference motion data is defined at the previous cartesian coordinate system, which means none of the 4 interested points provide a fixed origin. In this case, the difficulty of the motion retargeting process is significantly increased. However, compared to general robotic manipulators, they are commonly fixed in operational environments, which gives a fixed base origin. Therefore, a fixed base origin for raw reference motion is required to build the correspondence. Select $\mathbf{p_1}$ as fixed key point and the rests $\mathbf{p_i}= (x_i, y_i, z_i), i \in \{0, 1, 2, 3\}$ are transformed to a temporary frame $f_{temporaty}$ where $\mathbf{p_1}$ is temporarily fixed origin by $\mathbf{p_i} - \mathbf{p_1}$. After obtaining the temporarily fixed origin, a transformation $\mathbf{R}$ is applied to $\mathbf{p_0}$ to rotate $\overrightarrow{p_1p_0}_{base} = \mathbf{R}\cdot \mathbf{p_0}$ vertically to match the vertical robotic manipulator base. The resulting intermediate raw reference motion is shown in the middle part of \hyperref[fig:Fig/trans_ret.png]{Figure 3-5}\\

\[
\mathbf{R} = \begin{pmatrix}
\cos^2(\frac{\theta}{2}) + \sin^2(\frac{\theta}{2}) & 2(\sin(\frac{\theta}{2}) \cdot \sin(\frac{\theta}{2})) & 2(\sin(\frac{\theta}{2}) \cdot \cos(\frac{\theta}{2})) \\
2(\sin(\frac{\theta}{2}) \cdot \sin(\frac{\theta}{2})) & \cos^2(\frac{\theta}{2}) + \sin^2(\frac{\theta}{2}) & 2(\sin(\frac{\theta}{2}) \cdot \cos(\frac{\theta}{2})) \\
2(\sin(\frac{\theta}{2}) \cdot \cos(\frac{\theta}{2})) & 2(\sin(\frac{\theta}{2}) \cdot \sin(\frac{\theta}{2})) & \cos^2(\frac{\theta}{2}) + \sin^2(\frac{\theta}{2})
\end{pmatrix} \tag{1}
\]

\noindent \textbf{Link Rescaling.} Rescaling the link is to further correspond the raw reference motion and robotic manipulator. The scales used are taken from KUKA LBR iiwa 7 R800 datasheet as shown in \hyperref[Appendix:Fig/kuka_datasheet.pdf]{Appendix A-1}, where the robotic manipulator model is exactly the same one simulated in PyBullet environment. Connecting adjacent key points $\mathbf{p_i}$ and $\mathbf{p_{i+1}}$ forms three links $\overrightarrow{p_ip_{i+1}}$. Rescaling process calculates the unit vector $\widehat{\overrightarrow{p_ip_{i+1}}} = \frac{\overrightarrow{p_ip_{i+1}}}{\| \overrightarrow{p_ip_{i+1}} \|}$ resulting in four unit link vector $\widehat{\overrightarrow{p_ip_{i+1}}}$, $i \in \{0, 1, 2\}$. The unit link vectors are scaled with scaling factor $k_i$, $i \in \{0, 1, 2, 3\}$ where $k_i$ is obtained from \hyperref[Appendix:Fig/kuka_datasheet.pdf]{Appendix A-1}, and the results of rescaling process are reflected in the last part of \hyperref[fig:Fig/trans_ret.png]{Figure 3-5}.\\

\section{Generate Reference Motion}

\noindent \textbf{Define Reference Motion.} After base matching and link rescaling, the morphology of reference motion is fully retargeted to that of robotic manipulators. Therefore, the correspondence is established between the robotic manipulator and the reference motion to be imitated. However, the retargeted motion does not guarantee within the feasible workspace of the robotic manipulator. To address this issue, inverse kinematics is introduced to solve a feasible reference motion for a robotic manipulator under the constraints that minimize the discrepancy between retargeted motion and reference motion to be solved. The source retargeted motion in terms of joint cartesian coordinate at each frame is denoted by $\mathbf{x_i(t)}, i \in \{0, 1, 2, 3\}$ and the robotic manipulator reference motion $\mathbf{x_i(q_{j, t})}, j \in \{0, 1, 2, 3, 4, 5\}$ is determined by $\mathbf{q_{j, t}}$ indicating each joint $j$ angular position $q$ at each frame $t$. Therefore, the final output reference motion is represented by joint angular position $\mathbf{q_{j,t}}$.\\

\noindent \textbf{Solve Reference Motion.} With the aforementioned clear definition of reference motion, inverse kinematics can be solved as an optimization problem. The objective function $\Theta$ of this optimization problem iteratively solves for feasible joint angular positions $\mathbf{q_{j, t}}$ that follow the end-effector trajectory and source retargeted reference motion $\mathbf{x_i(t)}$.
\[\Theta = \arg\min_{\mathbf{q_{j, t}}} \begin{pmatrix} \displaystyle\sum_{i}\sum_{t} \|\mathbf{x_i(t)}-\mathbf{x_i(q_{j, t})}\|^2 + \displaystyle\sum_{j}\sum_{t}(\mathbf{q_{j, t}}-\mathbf{q_{j, t, ref}})^T \cdot (\mathbf{q_{j, t}}-\mathbf{q_{j, t, ref}}) \end{pmatrix} \tag{2}\]
As $\mathbf{q_{j, t}} \in [lower bound, upper bound]$, this joint range limit guaranteed the optimization results, i.e. the reference motion, within the feasible workspace.

\noindent \textbf{Interpolating \& Smoothing.} The resulting reference motion may not meet the required quantity to train a control policy. To address this issue, Cubic Spline is introduced to interpolate $\mathbf{q_{j, t}}$ to the desired volume. The Euclidian distance is calculated for each successive data $L_2 = \|\mathbf{q_{j, t+1}} - \mathbf{q_{j, t}} \|$. For each interval where $L_2 > 0.1$, around $\frac{L_2}{0.01}$ data is interpolated to increase the data volume and reduce the gap between two reference motions. Each interpolated point $S_i(q_{j, t})$ is obtained by solving a cubic polynomial equation subjected to interpolation $S_i(q_{j, t}^i)$, continuity $S_i(q_{j, t}^i) = S_{i-1}(q_{j, t}^i)$, and boundary condition $S''(q_{j, t}^0 \& q_{j, t}^n) = 0$ that gives $a_i$, $b_i$, $c_i$, and $d_i$.
\[
\begin{aligned}
&S_i(q_{j, t}) = a_i + b_i(q_{j, t} - q_{j, t}^i) + c_i(q_{j, t} - q_{j, t}^i)^2 + d_i(q_{j, t} - q_{j, t}^i)^3 \\
\text{s.t.} \quad &S_i(q_{j, t}^i) = S_i(q_{j, t}), \quad S_i(q_{j, t}^i) = S_{i-1}(q_{j, t}^i), \quad S''(q_{j, t}^0 \& q_{j, t}^n) = 0
\end{aligned} \tag{3}
\]
Most importantly, the reference motion is potentially subjected to sudden changes resulting in undesirable twitching motion. Therefore, with the prior knowledge of entire reference motion data $\mathbf{q_{j, t}}$, Locally Weighted Scatterplot Smoothing (LOWESS) is applied to the entire dataset to smooth the overall trend while preserving local features.
\[\widehat{q_i} = \frac{\sum_{j=1}^{n} w_{ij} q_i}{\sum_{j=1}^{n} w_{ij}} \text{ ,where } w_{ij} = \left\{ \begin{array}{ll} (1 - \left| \frac{q_j - q_i}{d} \right|^3)^3 &,\text{if } |q_j - q_i| < d \\ \quad \quad \quad 0 &,\text{otherwise} \end{array} \right. \tag{4}\]
$\widehat{q_i}$ is the $i^{-th}$ smoothed reference data for each joint $j$ at frame $t$. $w_{ij}$ is the weight of sample $j$ w.r.t. sample $i$. $d=0.05$ is the bandwidth of the local window. Eventually, the resulting reference motion generated in PyBullet is shown in \hyperref[fig:Fig/ref.png]{Figure 3-6}.\\

\insertFig{0.8}{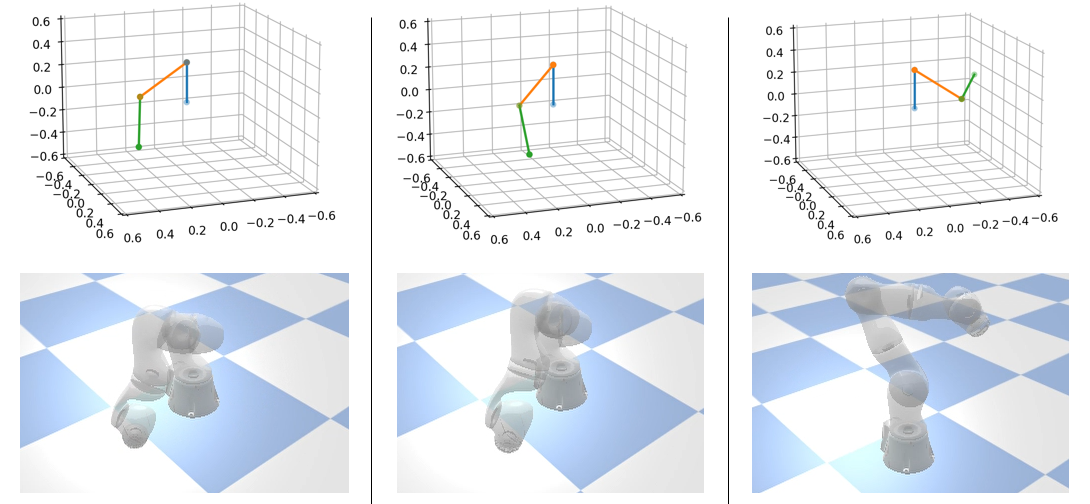}{Final Reference Motion Generated in PyBullet} 

\section{Formulate Reinforcement Learning Problem}

\noindent Motion Imitation can be explicitly defined as the process through which a robotic manipulator control policy is acquired, such that the control policy generates an action that imitates target reference motion. Therefore, the described process is naturally formulated as a Reinforcement Learning problem, where the objective is to obtain a policy $\pi$ that maximizes the designated expected return \Parencite{sutton2018reinforcement}.\\

\noindent \textbf{Feasibility Analysis.} The reference motion is essentially a trajectory defined in the time domain that consists of six joint’s $j$ angular position $\mathbf{q_{j, t}^{ref}}$ and absolute cartesian location $\mathbf{x_{j, t}^{ref}}$ at each timestep $t$. In the meanwhile, the actual motion performed by the robotic manipulator is represented in a similar format, $\mathbf{q_{j, t}^{rbt}}$ and $\mathbf{x_{j, t}^{rbt}}$. For each timestep $t$, the reference motion is defined independently from the other timestep. For a robotic manipulator, the motion at $t+1$ completely depends on the action $a_t$ executed in timestep $t$. Suppose each state $S_t$ is characterized or inferred by $\mathbf{q_{j, t}}$ and $\mathbf{x_{j, t}}$, future states $S_{t+1}$ completely depends on action $a_t$ executed by current states $S_t$. Therefore, motion imitation is a Markov Decision Process, i.e. $P(S_{t+1}|(S_{t}, a_{t}), (S_{t-1}, a_{t-1}), \ldots, (S_0, a_0)) = P(S_{t+1}|(S_{t}, a_t))$. This proved that formulating motion imitation as a Reinforcement Learning problem is feasible.\\

\noindent \textbf{Problem Formulation.} The robotic manipulator control agent (the agent), at each timestep $t$, receives state observation $\mathbf{s_t}$ from the environment and samples a continuous action $\mathbf{a_t} \sim \pi(\mathbf{a_t}|\mathbf{s_t})$ from its policy $\pi$. This sampled action $\mathbf{a_t}$ consists of six joints’ angular displacement $\mathbf{q_{j, t}^{rbt}}$. The agent applies this action $\mathbf{a_t}$, resulting in a new state observation $\mathbf{s_{t+1}}$ and a scaler reward signal $r_t$. Repeated interactions yields a trajectory $\tau = \{\ldots (\mathbf{s_t}, \mathbf{a_t}, \mathbf{s_{t+1}}), (\mathbf{s_{t+1}}, \mathbf{a_{t+1}}, \mathbf{s_{t+2}}), \ldots\}$ stored in experience replay buffer for updating the policy $\pi$. The agent aims to learn a control policy that maximizes the expected return with discount $\gamma$.
\[
\begin{aligned}
&J(\pi)=\mathbb{E}_{\tau \sim prob(\tau|\pi)}\left[\displaystyle\sum_{t=0}^{T-1}\gamma^tr_t\right] \quad \text{, where}\\
&prob(\tau|\pi)=prob(s_0) \displaystyle \mathbf{\prod_{t=0}^{T-1}}prob(\mathbf{s_{t+1}}|\mathbf{s_t}, \mathbf{a_t})\pi(\mathbf{a_t}|\mathbf{s_t}) 
\end{aligned} \tag{5}
\]
$prob(\tau|\pi)$ is the likelihood of a certain trajectory $\tau$ generated under policy $\pi$. $prob(\mathbf{s_0})$ is the probability of initial state observation being $\mathbf{s_0}$.

\noindent \textbf{Reward Design.} Motion Imitation reward $r_t$ is a scalar signal provided by the environment to the agent, which indicates the performance of the agent’s action $\mathbf{a_t}$ at a particular timestep $t$. Reward serves as a guide to the policy learning process by providing feedback on actions applied. Therefore, it is of great significance that design a reward suitable for the motion imitation task. Motion Imitation reward designed similar to Peng et. al. \Parencite{10.1145/3197517.3201311}.
\[
\begin{aligned}
&r_t = w^{p}r_{t}^{p} + w^{v}r_{t}^{v} + w^{e}r_{t}^{e} \quad \text{, where}\\
&w^{p} = -1.0, \quad w^{v} = -0.1, \quad w^{e} = -100.0
\end{aligned} \tag{6}
\]
\noindent The reward per timestep consists of three components, joints position reward $r_t^p$, joints velocity reward $r_t^v$, and end-effector position reward $r_t^e$, emphasizing imitation of reference motion in different domains. Each of the three rewards corresponds to a specific weight $w$, which indicates the level of influence of each reward on the final composite reward. All weights are negative meaning that at each timestep $t$. To maximize the overall expected return, it encourages the agent to sample an action $\mathbf{a_t}$ as close to the reference as possible.
\[
r_t^p = \displaystyle\sum_j coeff_j \cdot \mathbf{\|q_{jt}^{ref} - q_{jt}^{rbt}\|} \tag{7} \label{7}
\]
\noindent The joint position reward $r_t^p$ \eqref{7} reflects the absolute difference of joint positions. The $coeff_j$ indicates the importance level of each joint. As the most direct guidance of motion imitation, the corresponding weight $w^p$ is set to $w^p = -1$ indicating the ground level of influence.
\[
r_t^v = \displaystyle\sum_j coeff_j \cdot \mathbf{\|v_{jt}^{ref} - v_{jt}^{rbt}\|} \tag{8} \label{8}
\]
\noindent The velocity reward \eqref{8}, is the difference of two successive joint positions in unit time $\mathbf{v_{jt}} = \frac{\mathbf{q_{j,t}-\mathbf{q_{j,t-1}}}}{unitTime}$. Imitating this velocity gradient enables the agent to acquire prior knowledge of the next action $\mathbf{a_{t+1}}$ based on current action $\mathbf{a_t}$. As imitating velocity gradient is not a hard requirement but only to provide a tendency, the corresponding weight is set to $w^v=-0.1$.
\[
r_t^e = \mathbf{\|x_{t}^{ref} - x_{j,t}^{rbt}\|} \tag{9} \label{9}
\]
\noindent The most important requirement of motion imitation is to accurately track the end-effector position. Therefore, the corresponding weight is set to $w^e = -100$ to reflect this importance.

\noindent \textbf{Environment Design.} With the explicitly defined reward signal, the motion imitation task is completely formulated into a reinforcement learning problem. Although the problem is successfully formulated, the model without action and state space is yet to be completed. The action and state space should be constructed such that it serves the reward function to provide essential information to compute the reward signal. The desired action $\mathbf{a_t}$ for motion imitation task is six joint angular positions, $\mathbf{a_t} = \mathbf{q_{j, t}}$, where $j \in \{0, 1, 2, 3, 4, 5\}$. However, the model does not directly output the exact desired action, instead, it outputs a six-dimensional action space following a multivariate normal distribution. 
\[
\mathbf{a_t} \sim \mathcal{N}_6(\mu, \sigma) \quad \text{, where } \mu \in [-\pi, \pi] \tag{10}
\]
\noindent Sampling the exact joint angular position from a normal distribution guarantees the exploration of optimal imitating policy. The model outputs six mean values $\mu$ controlling overall sample tendency and the standard deviation $\sigma$ is shrinking over time to increase the sampling precision. But the randomness brought by multivariate normal distribution causes the action to oscillate, i.e. the randomness breaks the smoothness. Therefore, the final output is smoothed using exponential smoothing, where the smoothing parameter $\beta=0.03$. This $\beta$ put little emphasis on the most recent action, given that the joint angle difference between reference motion is also negligible.
\[
\mathbf{\hat{a}_{t+1}} = \beta \mathbf{a_t} + (1-\beta) \mathbf{\hat{a}_t} \tag{11}
\]
\noindent The state space $\mathbf{s_t}$ should be constructed such that it contains all normalized necessary observations of the motion imitation task. Although we have the reference motion made available prior to the agent training \& testing, to ensure consistency, the joint angular positions of both reference $\mathbf{q_{j,t}^{ref}}$ and robot $\mathbf{q_{j,t}^{rbt}}$ are retrieved from PyBullet simulation environment directly. With the observed joint angular positions, the position $\mathbf{q^{diff}}=\mathbf{q^{ref}} - \mathbf{q^{rbt}}$ and velocity difference $\mathbf{\omega^{diff}} = (\mathbf{q_{j, t}}-\mathbf{q_{j, t-1}})^{ref} - (\mathbf{q_{j, t}}-\mathbf{q_{j, t-1}})^{rbt}$ are computed to provide more comprehensive explanation to the observed information. Finally, end-effector Euclidian positions of both reference and robot are obtained to compute end-effector positional difference $\mathbf{x^{diff}} = \mathbf{x_t^{ref}-\mathbf{x_t^{rbt}}}$.
\[
\mathbf{s_t} = \{\mathbf{q_{j,t}^{ref}},\quad\mathbf{q^{diff}},\quad\mathbf{\omega^{diff}},\quad\mathbf{x^{diff}}\}\tag{12}
\]
\noindent The agent receives an observation state $\mathbf{s_t}$ at each timestep, enabling the agent to understand the immediate consequence of the action taken.

\noindent \textbf{Experience Sampling.} Interactions of every timestep are normalized and then stored in the experience replay buffer (the buffer). Especially for rewards, they are discounted $R_t = \sum_{i=t}^{T} \gamma^{i-t} \cdot r_i$ with $gamma=0.99$ focusing more on the recent rewards. When performing policy updates, a re-sampling process from the buffer de-correlates the consecutive samples from continuous trajectories generated by policy $\pi$ hence improving the efficiency of data.\\
\[
R_t = \sum_{i=t}^{T} \gamma^{i-t} \cdot r_i \tag{13}
\]

\noindent \textbf{Objective Function.} The overall objective function consists of three parts, the clipped surrogate loss (Surr Loss), negative mean squared error loss (MSE Loss), and entropy loss \parencite{zhang2020proximal}. Surr Loss reflects the policy’s quality in motion imitation tasks, MSE Loss evaluates state values improving state value estimation accuracy, and entropy loss encourages the policy’s exploration. Maximizing this objective function is the primary goal of the formulated reinforcement learning problem.
\[
J(\theta)=L_{\text{policy,clip}}(\theta) + w_{\text{v,MSE}}\cdot L_{\text{value,MSE}}(v, R) + w_{\text{entropy}}\cdot H(\pi_{\theta}) \tag{14}
\]
\noindent Surr Loss \eqref{15} is derived from current $\pi_{\theta}(\mathbf{a_t|s_t})$ and previous $\pi_{\theta-1}(\mathbf{a_t|s_t})$ policy. It evaluates the level of policy improvements. The ratio between the probability of actions performed by the current policy and that of the previous policy $\frac{\pi_{\theta-1}(a_t|s_t)}{\pi_{\theta}(a_t|s_t)}$ reflects the improvements, either positive or negative, of the $\pi_{\theta}$ over $\pi_{\theta-1}$. The advantage $A^{\pi_{\theta-1}}(s_t, a_t)$ brought by the current policy compared to the previous policy serves as the weight of Surr Loss. The overall Surr Loss $\frac{\pi_{\theta-1}(a_t|s_t)}{\pi_{\theta}(a_t|s_t)}\cdot A^{\pi_{\theta-1}}(s_t, a_t)$ is clipped by $\epsilon=0.2$ to limit the speed of policy update.
\[
L_{\text{policy,clip}}(\theta) = \mathbb{E}_t \left[ \min \left( \frac{\pi_{\theta-1}(a_t|s_t)}{\pi_{\theta}(a_t|s_t)} A^{\pi_{\theta-1}}(s_t, a_t), \text{clip}(\epsilon, 1-\epsilon) \right) \right] \tag{15} \label{15}
\]
\noindent MSE Loss \eqref{16} computes the mean squared error between state values $v_t$ and discounted rewards $R_t$, where state values are predicted values of states and discounted rewards are actual values of states. The smaller MSE Loss leads to more accurate state evaluation, further improving policy updates. Therefore, to align with the goal of maximizing the objective function, the negative MSE Loss weight $w_{\text{v,MSE}}=-0.5$ is applied to the overall objective function.
\[
L_{\text{value,MSE}}(v, R) = \frac{1}{n} \displaystyle\sum_{t=1}^{n} (v_t - R_t)^2 \tag{16} \label{16}
\]
\noindent Entropy Loss \eqref{17} evaluates the entropy of the policy $\pi_{\theta}$ distribution at a specific state $s_t$. It refers to the uncertainty or randomness of the policy’s action probabilities. A higher entropy loss implies that the policy is more exploratory, improving the agent’s ability to explore the environment comprehensively and thus converging to a better policy. For continuous action space under normal distribution $\mathbf{a_t} \sim \mathcal{N}_6(\mu, \sigma)$, the entropy loss is equivalent to the entropy of this normal distribution. However, unlimited exploration causes slower convergence and brings instability. Therefore, the weight $w_{entropy}= 0.01$ corresponding to entropy loss tends to be small.
\[
H(\pi_{\theta}) = \frac{1}{2}\log(2\pi_{\theta} e \sigma^2) \tag{17} \label{17}
\]

\section{Motion Imitation}

\noindent The Motion Imitation is fully formulated as reinforcement learning problem and the complete process is shown in \hyperref[fig:Fig/moim_init.png]{Figure 3-7}. Through the repeated interaction between the agent and the environment, the policy is updated iteration by iteration until converges to near optimal.

\insertFig{1.0}{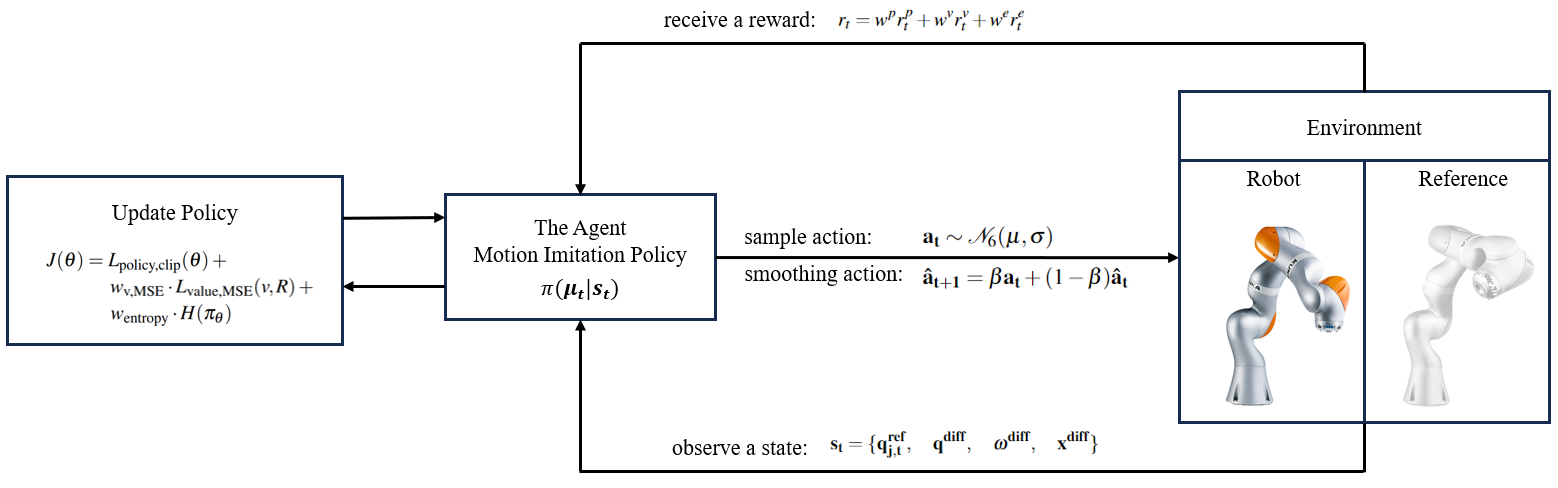}{Motion Imitation task formulated as a reinforcement learning problem} 

\section{Build PyBullet Simulation Environment} 

\noindent PyBullet provides an excellent physics simulation environment for Motion Imitation. The aforementioned reinforcement learning problem is constructed in PyBullet simulation environment with the provided API. Physical properties related to the robotic manipulator such as kinematic, dynamic, and inertia are simulated to match that of the real-world scenarios. However, for reference motion, all relevant kinematic, dynamic, and inertia properties are disabled, making the reference motion an empty shell for visualization purposes.
\newpage

\lhead{Experiments}
\chapter{Experiments}

\section{Experiments Objective}

\noindent The main objective of this experiment is to acquire the motion imitation control policy (the policy) from an input reference video, and generalize the policy to imitate unseen arm motion from any inference videos hence evaluating the policy performance. The learned motion imitation policy is then evaluated by different metrics on inference videos to reflect the policy performance.\\

\section{Experiments Setup}

\noindent \textbf{Input Videos Requirements.} For the policy learning process, a wide variety of reference motions should be provided, allowing the agent to adapt to imitation across different positions and orientations. Hence the reference video should contain as complex and diverse arm motion as possible that enables the agent to accumulate relevant experience during the training process, and the accumulated experience can greatly benefit the generalization ability to unseen cases. On the other hand, inferencing the learned policy does not require complex video inputs, however, it is preferred with meaningful smooth arm motion. As the objective of motion imitation is to imitate human arm motion and replicate the respective function of each motion, the input videos should focus more on arm motion rather than whole-body movements. However, existing human pose datasets such as Human3.6M \Parencite{ionescu2013human3} and Human-eva \Parencite{sigal2010humaneva} are for human pose estimation, which emphasizes more on whole-body movements. In contrast, arm motions in ballet dance videos are smooth but complex enough to cover different types of arm motion such as lifting, circulating, waving, etc., therefore, they are selected as sourcec of expert reference motion. Additionally, these motions are frequently performed by production line operators, thereby benefiting the practical implementation of substituting human operators with robotic manipulators using recorded videos of these operators.\\

\noindent \textbf{Environments \& Hardwares.} Motion imitation processes are simulated in PyBullet \Parencite{coumans2019pybullet} environments using kuka iiwa robotic manipulator \Parencite{kuka-iiwa} in \hyperref[sec:appendix_section]{Appendix A.2}. The expert reference motion used in the training process is extracted from a ballet dance video and the reference motion used in the evaluation are videos of daily common motion, production line human operator, and warehouse pick and place. All experiments were conducted remotely in Conda virtual env named “MoIm” on a virtual machine with Linux Ubuntu 20.04 LTS CUDA version 12.2 with GPU Nvidia RTX 4080Ti. Different hyperparameter settings were applied in the experiments to optimize performance and enhance generalization abilities.\\

\noindent \textbf{Evaluation Metrics.} Similarity is evaluated by summing the difference of angular positions $\mathbf{ q_{j, t}^{diff}=q_{j, t}^{ref}-q_{j, t}^{rbt}}$.
\[
\delta_{\text{similarity}} = \sum_{j=1}^{n} q_{j, t}^{diff} \tag{18}
\]
\noindent End-Eff is evaluated by directly computing the Euclidean distance between reference $\mathbf{x^{ref}}$ and robot $\mathbf{x^{rbt}}$ end-effector position.
\[
\delta_{\text{end-eff}} = \|\mathbf{x_i^{ref}} - \mathbf{x_i^{rbt}}\|_{l2}^2 \tag{19}
\]
\noindent However, considering the small joint angular position difference may accumulate such that the overall deviation is large. To address this issue, the mean per joint position error is introduced to evaluate the average Euclidean distance between reference and robot motion \Parencite{fang2018learning}.
\[
\delta_{\text{MPJPE}} = \frac{1}{N} \sum_{i=1}^{N} \| \mathbf{x_{j,t}^{ref}} - \mathbf{x_{j,t}^{rbt}} \| \tag{20}
\]
\noindent $\delta_{\text{similarity}}$ directly evaluates the difference between output action (joint angular positions) to reference motion, which is the most direct metric that indicates the extent of imitation. $\delta_{\text{end-eff}}$ is a metric representing whether the robot end-effector can follow that of the reference. It is the most important metric in the task of motion imitation failing which the end-effector no longer replicates the desired reference trajectory, hence may lead to the failure of the task to be imitated. $\delta_{\text{MPJPE}}$ addresses the accumulated error issue, which evaluates whether each individual joints of the robot coincide with that of reference motion.

\section{Implementation Details}

\noindent \textbf{Simulation Environment} The experiments including policy learning and policy evaluation are done via PyBullet simulation environment with open-source kuka iiwa unified robotic description format (urdf). The robot base position and orientation are set to $[-0.25, 0.0, 0.6]$ and $[0, 0, 0, 1]$, which places the kuka iiwa above the simulated plane facing the positive x-axis. There are no specific reasons for this setup except for better visualization purposes. Table \ref{tab:kuka_dynamic} lists the dynamics detailed information of kuka iiwa in simulation, where $\mu$ refers to the overall friction coefficient as well as damping effect and $\epsilon$ refers to restitution coefficient.\\

\begin{table}[htbp]
\centering
\begin{tabular}{c|c|c|c|c}
\hline
Index & Mass ($kg$) & Inertia ($kg \cdot m^2$) & $\mu_{\text{overall damping}}$ & $\epsilon$ \\
\hline
$j_0$ & 0.5 & 4.0 & -1 & 1e-3\\
$j_1$ & 0.5 & 4.0 & -1 & 1e-3\\
$j_2$ & 0.5 & 3.0 & -1 & 1e-3\\
$j_3$ & 0.5 & 2.7 & -1 & 1e-3\\
$j_4$ & 0.5 & 1.7 & -1 & 1e-3\\
$j_5$ & 0.5 & 1.8 & -1 & 1e-3\\
$j_6$ & 0.5 & 0.3 & -1 & 1e-3\\
\hline
\end{tabular}
\caption{Dynamic parameters simulated in PyBullet kuka iiwa}
\label{tab:kuka_dynamic}
\end{table}
\noindent However, reference kuka only serves as a visualization aid, therefore, all dynamics properties related to reference kuka are disabled for convenience. Other than reference and robot parameters, the global PyBullet simulation timestep is set to $\frac{1.}{240.}$ s and the gravitational acceleration is set to $(0, 0, -9.8)$ $m \cdot s^{-2}$, where negative indicates pointing to negative z-axis.

\insertFig{0.9}{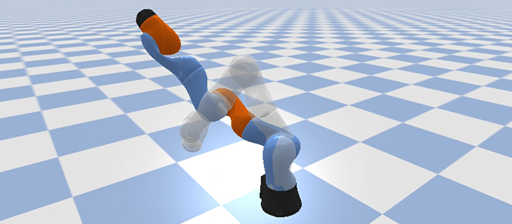}{PyBullet simulation environment, reference(shadow), robot(color)}

\noindent The final PyBullet simulation environment are shown in \hyperref[fig:Fig/kuka_sim.png]{Figure 4-1}. Both reference (shadow) and robot (color) are levitated above the ground in case of collision with it. The reference model only serves as a visualization purpose indicating what is the desired or target motion at each timestep. The imitation policy controls only the robot model to imitate reference motion.\\

\noindent \textbf{Acquire Reference.} The reference motions are generated by solving inverse kinematics as a constrained optimization problem as illustrated in the methodology section. PyBullet provides an inverse kinematics iterative solver. Therefore, the constrained optimization problem is solved by feeding the endpoint trajectory of the skeletal arm structure as the desired goal and the rests key points positions as constraints. Table \ref{tab:ik_constraints} lists the constraints adopted for the experiments.\\

\begin{table}[htbp]
\centering
\begin{tabular}{c|c|c|c|c|c}
\hline
Index & lower-limit (rad) & upper-limit (rad) & joint-range (rad) & joint-damping & rest-pose\\
\hline
$j_0$ & -2.96 & +2.96 & 5.8 & 0.1 & $q_{0}^{ref}$\\
$j_1$ & -2.09 & +2.09 & 4.0 & 0.1 & $q_{1}^{ref}$\\
$j_2$ & -2.96 & +2.96 & 5.8 & 0.1 & $q_{2}^{ref}$\\
$j_3$ & -2.09 & +2.09 & 4.0 & 0.1 & $q_{3}^{ref}$\\
$j_4$ & -2.96 & +2.96 & 5.8 & 0.1 & $q_{4}^{ref}$\\
$j_5$ & -2.09 & +2.09 & 4.0 & 0.1 & $q_{5}^{ref}$\\
$j_6$ & -3.05 & +3.05 & 6.0 & 0.1 & $q_{6}^{ref}$\\
\hline
\end{tabular}
\caption{PyBullet inverse kinematics null-space constraints}
\label{tab:ik_constraints}
\end{table}

\noindent These constraints guarantee that the reference motions are generated within the feasible workspace. Most importantly, by setting rest pose at reference motion positions, the inverse kinematic (ik) solver provides the solution that is closest to reference motion at the meanwhile satisfying all other constraints. However, how to acquire the reference motion prior to the ik solution? So long as the input video is smooth, the positional difference between two consecutive arm skeletal motions is negligible. Therefore, the reference motion needed at the current timestep $t$ can be approximated with the reference motion at the previous timestep $t-1$. As for the initial reference motion, it is calculated by solving regular inverse kinematics equations taking into account the Cartesian coordinate of each joint, without involving PyBullet iterative ik solver.\\

\noindent \textbf{Policy Learning.} The motion imitation policy is represented by proximal policy optimization (ppo) network parameters and the network architecture is shown in table \ref{tab:net_arch}. Additionally, all hyperparameters related to ppo imitation policy learning experiment are listed in Table \ref{tab:hyper_params}.\\

\begin{table}[htbp]
\centering
\begin{tabular}{l|l|l|l}
\hline
Components  & Layer Type        & Output Shape  & Parameters\\
\hline
            & Linear(21, 256)   & (, 256)       & 5376+256\\
Actor       & Tanh()            & (, 256)       & 0\\
            & Linear(256, 256)  & (, 256)       & 65536+256\\
            & Tanh()            & (, 256)       & 0\\
            & Linear(256, 6)    & (, 6)         & 1536+6\\
            & Tanh()            & (, 6)         & 0\\
\hline
            & Linear(21, 256)   & (, 256)       & 5376+256\\
Critic      & Tanh()            & (, 256)       & 0\\
            & Linear(256, 256)  & (, 256)       & 65536+256\\
            & Tanh()            & (, 256)       & 0\\
            & Linear(256, 1)    & (, 1)         & 1536+1\\
\hline
\end{tabular}
\caption{Policy network architecture and parameters}
\label{tab:net_arch}
\end{table}

\noindent Tanh function is selected as activation mainly for controlling kuka iiwa to perform both positive and negative rotation, the activated output $a_t \in [-1, 1]$ is linearly remapped to $a_t \in [-\pi, \pi]$.\\

\noindent \textbf{Policy Evaluation.} The obtained motion imitation policy (the policy) is evaluated with the aforementioned metrics, i.e. $\delta_{similarity}$, $\delta_{end-eff}$, and $\delta_{MPJPE}$. At the same time, to evaluate the generalization ability of the policy, reference motions used in the evaluation process consist of arm motion packaging, eating, and random waving generated from inference videos specifically reserved for evaluation (unseen data). Finally, the learning process itself is reflected in an accumulated reward curve to indicate the performance of the policy learning process.\\

\newpage

\lhead{Results and Discussions}
\chapter{Results and Discussions}

\section{Policy Learning}

\noindent The policy learning process requires approximately 1.5 hours. Overall, this curve reflects a typical process of policy learning, where the agent quickly comprehends the unfamiliar environment at the beginning and then grasps general task objectives. Subsequently, it gradually optimizes its policy until convergence.

\insertFig{0.9}{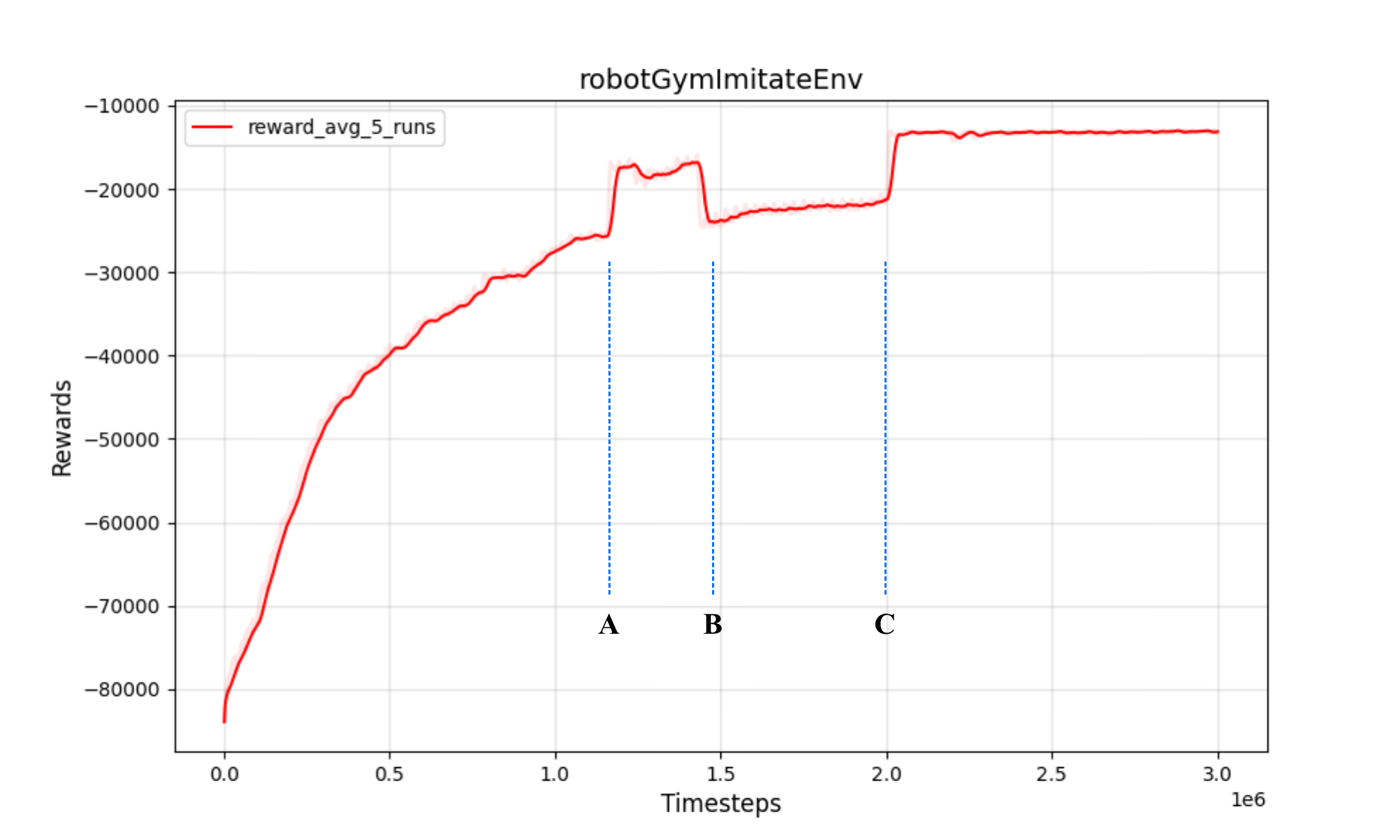}{Average reward (5 runs) of the policy learning process}

\noindent However, there are three noticeable sharp increases or decreases as indicated by A, B, and C on \hyperref[fig:Fig/learning_curve.png]{Figure 5-1}. Before sharp increase A, as the weight of the end effector position reward $w^{e}=100$, the agent attempts to output actions that strictly imitate the end effector. At sharp increase A, the agent attempts to output actions that imitate end effector position and joint angles, hence it creates a sharp increase in reward. At sharp decrease B, the agent has shifted its attention back to imitating the end effector position. However, at sharp increase C, the agent eventually strikes a balance between imitating the end effector position and joint angles. Therefore, after sharp increase C, the policy converges and has no further improvements.

\section{Policy Evaluation}

\noindent The learned policy is evaluated using three video inputs. The first one “packaging” mainly focuses on random left and right motions, the second one “eating” mainly focuses on random up and down motions, and the last one “waving” mainly focuses on circular or curved motions. These three videos contain the common motions (left right, up down, curve) in daily arm motions, implying the feasibility of replacing human workers with robotic manipulators in special tasks. For convenience, Table \ref{tab:videos} shows the evaluation videos along with their aliases.\\
\begin{table}[htbp]
\centering
\begin{tabular}{l|l|l|l}
\hline
video name & motion evaluated & video alias & frames\\
\hline
packaging & random left and right & $Video.LR $ & 406\\
eating & random up and down & $Video.UD $ & 431\\
waving & circular or curved & $Video.CC $ & 737\\
\hline
\end{tabular}
\caption{Evaluation videos reference alias and frame number} 
\label{tab:videos}
\end{table}

\noindent \hyperref[fig:Fig/demo_1.png]{Figure 5-2} selected several typical motions evaluated including simple moving (i-ii) and complex folding (iii-vi). As is obviously shown in \hyperref[fig:Fig/demo_1.png]{Figure 5-2}, the learned policy succeeded in imitating both motions but it deviated from reference motion when imitating complex folding. Detailed evaluation results are illustrated in the following discussions.

\insertFig{0.9}{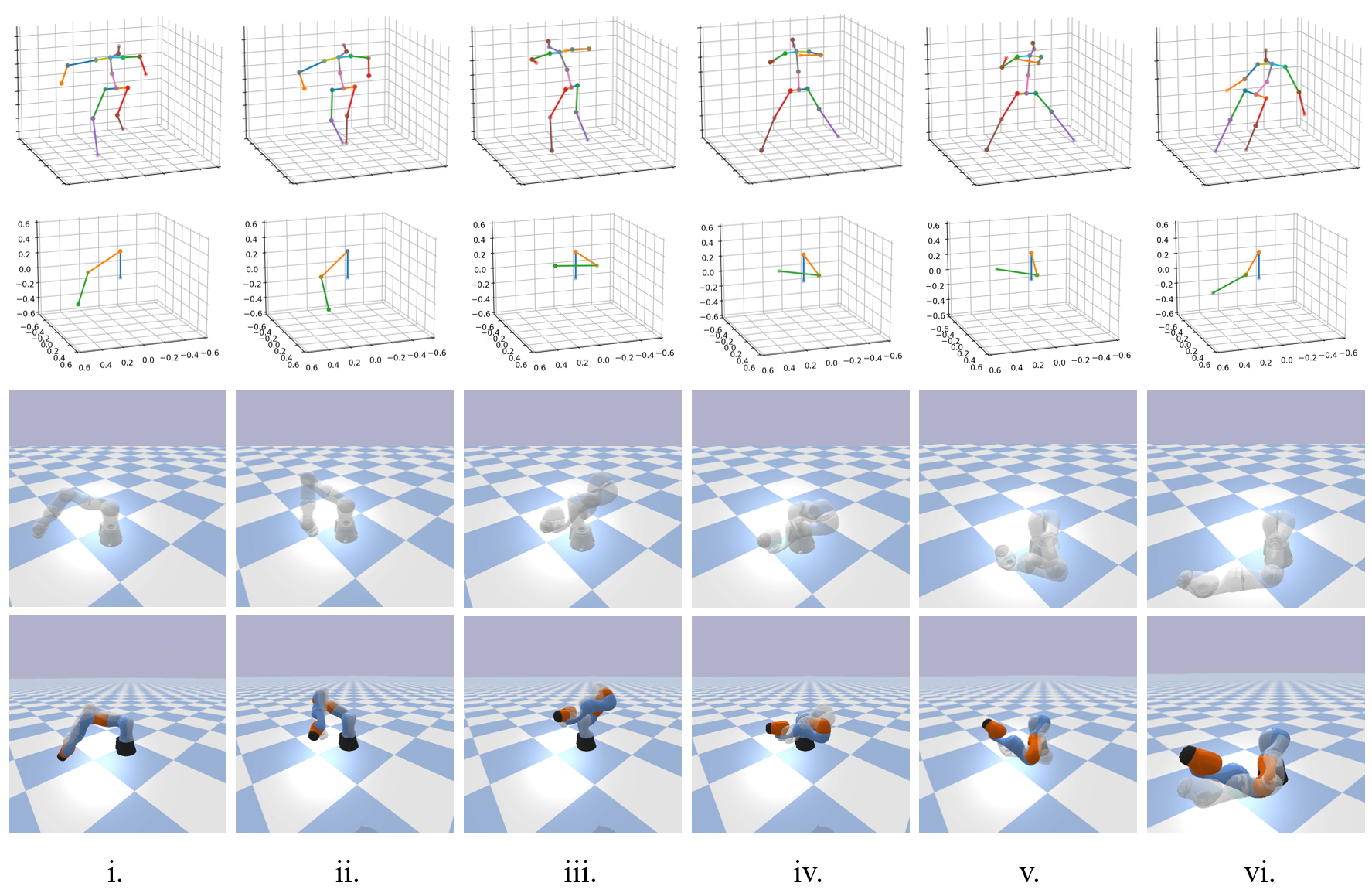}{Visualization of typical motions evaluated}

\noindent The evaluation results in terms of the aforementioned metrics are shown in Table \ref{tab:eva_results}. The learned policy excelled in imitating horizontal left and right motions, which resulted in approximately 0.09 m error in the following end effector and 0.056 m error in imitating each joint's position. However, the learned policy poorly behaved in imitating vertical up and down motions, as it shows the largest error for all the metrics. Finally, the learned policy achieved a rather satisfying result in imitating circular or curved motions, which gives 0.094 m error in imitating each joint's position. But circular motions as a composite of both horizontal and vertical motions, the learned policy inherited poor behaviors in imitating vertical motions, therefore resulting in a relatively large error in imitating the end effector position.\\
\begin{table}[htbp]
\centering
\begin{tabular}{c|c|c|c|c}
\hline
video & ref motions & $\delta_{similarity} (rad)$ & $\delta_{end-eff} (m)$ & $\delta_{MPJPE} (m)$\\
\hline
$video.LR$ & (2648, 6) & -0.22087 & \textbf{0.09120} & \textbf{0.05664}\\
$video.UD$ & (4898, 6) & -0.28610 & 0.26427 & 0.16258\\
$video.CC$ & (4748, 6) & \textbf{0.16300} & 0.19572 & 0.09400\\
\hline
\end{tabular}
\caption{Evaluation results evaluated by metrics $\delta_{similarity}$, $\delta_{end-eff}$, and $\delta_{MPJPE}$} 
\label{tab:eva_results}
\end{table}

\insertFig{0.9}{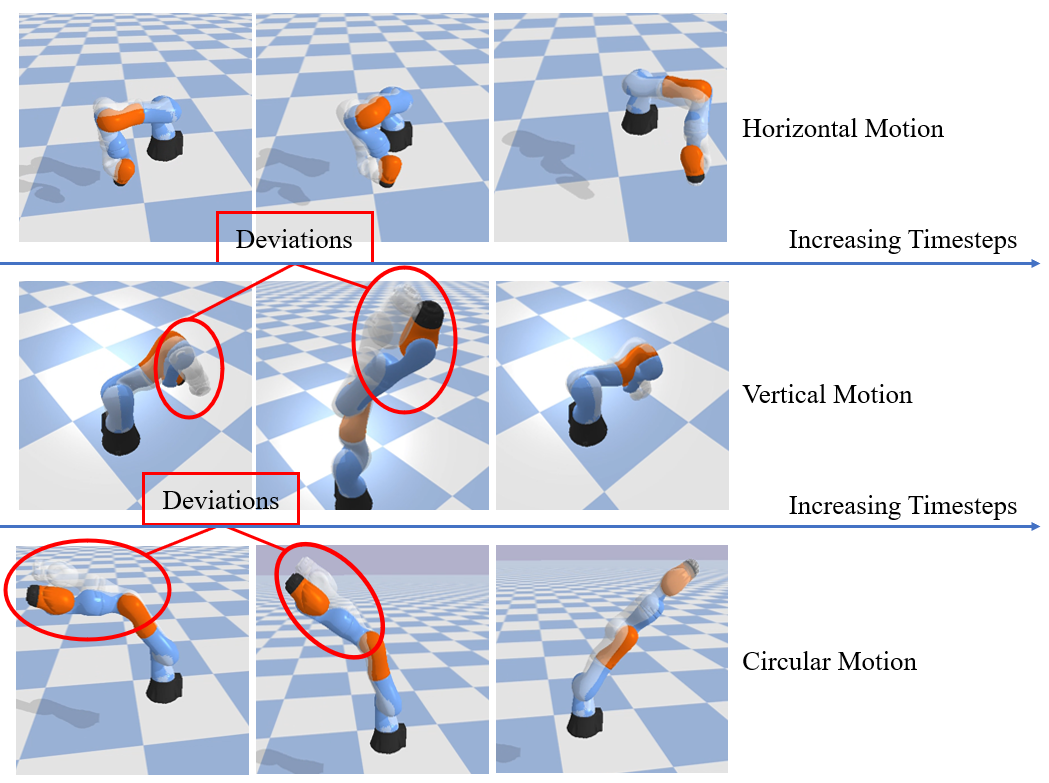}{The deviations of the learned policy corresponding to evaluation metrics}

\noindent As shown in \hyperref[fig:Fig/deviations.png]{Figure 5-3}, the motion imitation model has acquired a policy capable of replicating reference motion generally. However, \hyperref[fig:Fig/deviations.png]{Figure 5-3} also illustrates the corresponding evaluation metric results, indicating notable deviations in vertical and circular motions highlighted by red circles. These deviations commonly occur during the transition between points as time progresses. This phenomenon arises because the 3D HPE used in the skeletal motion extraction process is not sensitive to vertical arm motions. Consequently, only sharp changes in human motion lead to noticeable vertical skeletal motions. Such sharp changes cause delays in policy reactions, particularly when the vertical reference motions involve relatively high speeds. Therefore, the vertical movement process deviates from the reference motions.\\

\insertFig{0.9}{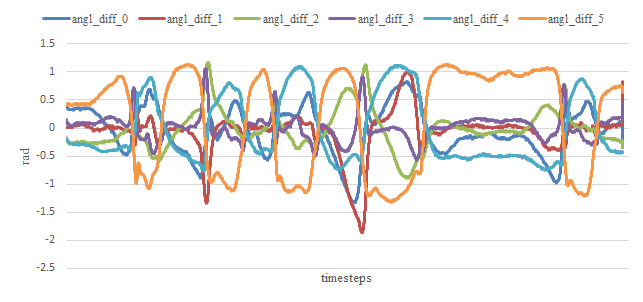}{Joint angle differences between the reference and robot motions ($video.CC$)}
\insertFig{0.9}{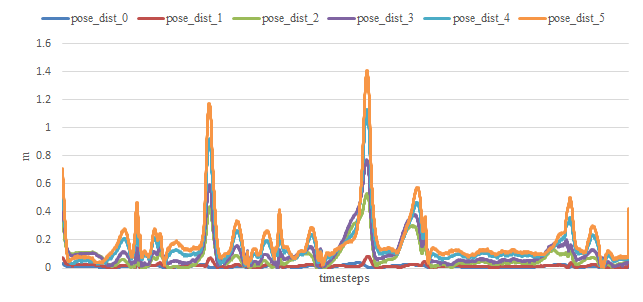}{Link position difference between the reference and robot motions ($video.CC$)}

\noindent \hyperref[fig:Fig/cc_ang.png]{Figure 5-4} illustrates the differences in joint angles, while \hyperref[fig:Fig/cc_pos.png]{Figure 5-5} illustrates the differences in link positions in comparison to the reference motions. All differences reflected in the figures are recorded in one episode of imitation, i.e. complete imitation of all reference motions. Reference motions used are from $video.CC$ as it contains both horizontal and vertical motions hence more representative. The “one episode differences curve” of $video.LR$ and $video.UD$  can be found in \hyperref[sec:appendix_ep]{Appendix A.3}.\\

\noindent \textbf{Episodic Angle Differences.} The episodic joint angle difference obviously follows a periodic fluctuation trend. This periodic behavior arises from the policy learning process. In the process of policy learning, actions that result in significant deviations are discouraged. Moreover, the agent attempts to produce the opposite actions in the next output to seek higher rewards. While effective initially, this behavior leads to a cycle of alternating actions until even the opposite actions are no longer encouraged, thereby giving rise to periodic fluctuations. This periodical compensation doesn't solely occur within the angles of one joint. Instead, the angles of various joints compensate each other to optimize overall imitation. This is well demonstrated in \hyperref[fig:Fig/cc_ang.png]{Figure 5-4}, where when certain joint angles deviate, the angles of other joints strive to compensate, resulting in out-of-phase periodic fluctuations. This compensation behavior finally results in lower $\delta_{similarity}$.\\

\noindent \textbf{Episodic Link Differences.} The episodic link position difference also shows a fluctuation without any period trend. The intensity of fluctuation increases as the link position moves from inner (dark blue, first link) to outer (orange, end effector) as the accumulated position errors are eventually reflected at the most outer link, i.e. end effector. However, \hyperref[fig:Fig/cc_pos.png]{Figure 5-5} perfectly align with the aforementioned behavior in \hyperref[fig:Fig/deviations.png]{Figure 5-3}: large deviations occur in the transition from one point to another. Each peak resents a deviation, but regardless of their magnitude, these deviations dissipate after several timesteps as the reference motion and the end effector of the robot converge. I.e., while the policy may not match the speed of the moving end effector trajectory, it eventually manages to imitate the trajectory itself. Although each peak leads to an increase in $\delta_{MPJPE}$, the averaged value is still within a satisfactory range. Moreover, even though $\delta_{end-eff}$ is relatively large in $video.UD$ and $video.CC$ evaluations, it might be the end effector is in the transition of two steady motions.\\

\section{Discussion and Future Work}

\noindent \textbf{Discussions of Existing Model.} The present motion imitation model provides only a baseline of imitating human arm motions from given videos. Overall, the motion imitation model (the model) is able to learn a desired imitating control policy that can successfully imitate the target motions. Given the fact that the model only takes a few seconds of video as input yet produces a motion-imitating policy capable of generalizing well to other arm motions from unseen videos, it simplifies the expert data-obtaining process and excels in the task of imitation.\\

\vspace{1cm}
\insertFig{0.9}{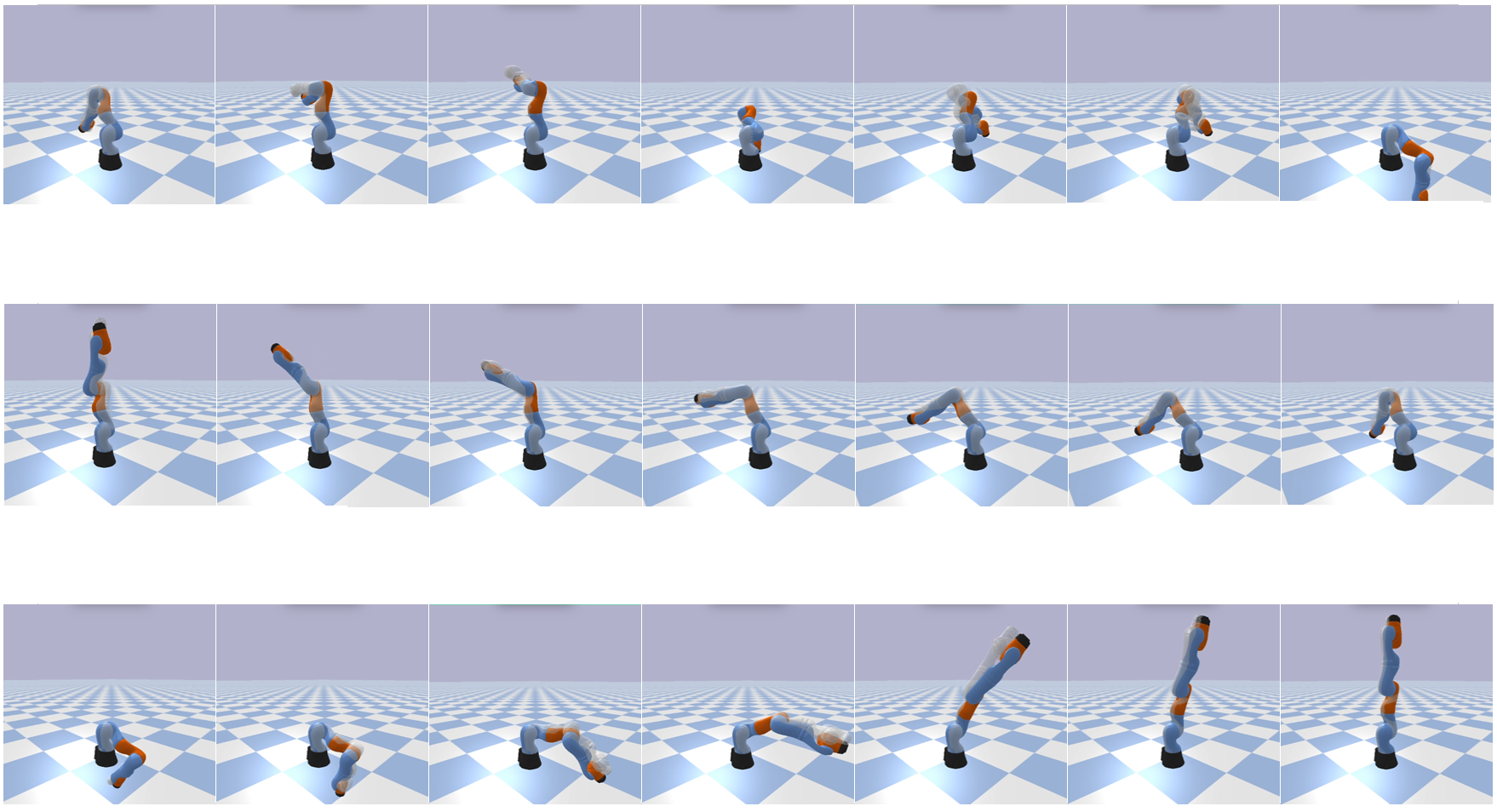}{The results of the motion imitation model}
\vspace{1cm}

\noindent \hyperref[fig:Fig/moim.png]{Figure 5-6}illustrates the performance of the Motion Imitation model. Overall, the model is capable of learning an excellent policy to imitate reference motions. However, the policy exhibits a slight delay in responding to rapid changes in the reference motions, resulting in delayed imitation of fast movements. Despite this delay in imitation, the policy learned by the model is still able to replicate trajectories. This delay may partially affect the real-time nature of the motions and reduce evaluation metric scores, but overall, the model can still successfully imitate reference motions.\\

\noindent \textbf{Limitations and Future Works.} However, the limitations are also obvious. Although the model learns an imitating policy based on a short video, the quality of the policy indeed heavily depends on the input videos. Therefore, effort is still required to find or self-recording a suitable input video. In the process of motion extractions, a better 3D HPE model leads to more accurate extracted motion, hence influencing on overall motion imitation process. Moreover, in motion retargeting, an alternative way of solving inverse kinematics may be able to generate better reference motions further improving the imitation. Most importantly, the present model only achieved basic motion imitation, more complex tasks such as imitating a series of motions while grasping the objects placed near the trajectory. This can be done by designing task-specific environments and rewards and formulating a more complex and challenging reinforcement learning problem. Table \ref{tab:future} summarises potential limitations and possible future work for convenience.\\
\begin{table}[htbp]
\centering
\begin{tabular}{c|c}
\hline
Limitations & Possible Future Works\\
\hline
Training Video & Better training video covering more complex motions.\\
3D HPE & Improving 3D HPE accuracy.\\
Motion Retarget & Alternative ways of solving ik and retargetings.\\
Reinforcement Learning & Formulate more complex and challenging RL problem.\\
\hline
\end{tabular}
\caption{Summarise limitations and possible future works} 
\label{tab:future}
\end{table}

\lhead{Conclusions}
\chapter{Conclusion}

\section{Project Summary}

\noindent This project combines reinforcement learning with 3D human pose estimation, providing a novel Motion Imitation model that excels at imitating the motions of any human arm in input videos. Moreover, unlike traditional imitation learning which requires a large amount of expert data for policy learning, the Motion Imitation model simplifies the complex motion imitation process into a robotic manipulator joint angle prediction problem. This means that only a few seconds of video covering complex arm motions are needed to generate a small amount of expert data, which can then be used to learn the policy for motion imitation. Furthermore, the policy produced by this model exhibits strong generalizability, allowing learned techniques to be easily transferred to imitate arm motions generated from unfamiliar videos. The final evaluation results demonstrate that the learned strategies of the model effectively mimic the target actions, particularly in replicating the positions of each link and the end effector. In summary, this project contributes a lightweight, convenient, user-friendly, and highly accurate model to the field of Motion Imitation.\\

\section{Potential Contribution}

\noindent This model provides a novel approach to Motion Imitation, enabling relatively good results with minimal expert data. Furthermore, it demonstrates relatively good accuracy in imitating motions such as swaying left and right, swaying up and down, and curved motions, which are common among production line workers. Therefore, the model has the potential to be used to imitate their motions based on videos, further achieving the goal of replacing them. Additionally, the model serves as a baseline, suggesting the possibility of designing more complex reinforcement learning problems in the future. These problems could allow policies to perform tasks such as grasping and placing objects based on the foundation of imitating human arm movements.

\newpage

\addcontentsline{toc}{chapter}{References}
\printbibliography[title={References}]

  \clearpage
  \pagenumbering{arabic}
  \renewcommand{\thepage}{R-\arabic{page}}
\lhead{}

\newpage
\appendix
\renewcommand{\chapname}{Appendix}
\pretocmd{\chapter}{
  \clearpage
  \pagenumbering{arabic}
  \renewcommand*{\thepage}{\thechapter-\arabic{page}}
}{}{}
\rhead{Appendix}

\chapter{Appendix}

\section{Hyperparameters used in Motion Imitation Model}

\begin{table}[htbp]
\centering
\begin{tabular}{l|l|l}
\hline
hyperparameters             & Descriptions              & Values\\
\hline
$E$                         & training epochs           & 30\\
$T_{\text{max-episode}}$    & max episode length        & 1e3\\
$T_{\text{max}}$            & max training steps        & 2e6\\
$\epsilon_{\text{clip}}$    & epsilon clipping          & 0.2\\
$\gamma$                    & discount factor           & 0.99\\
$\beta$                     & entropy loss weight       & 0.01\\
$\alpha_{\text{actor}}$     & actor learning rate       & 0.0003\\
$\alpha_{\text{critic}}$    & critic learning rate      & 0.001\\
$\sigma$                    & action standard deviation & 0.6\\
$\sigma_{\text{decay}}$     & action std decay rate     & 0.05\\
$\sigma_{\text{min}}$       & min action std            & 0.1\\
\hline
\end{tabular}
\caption{Policy network architecture and parameters}
\label{tab:hyper_params}
\end{table}

\section{KUKA LBR iiwa 7 R800}
\label{sec:appendix_section}

\insertFig{0.9}{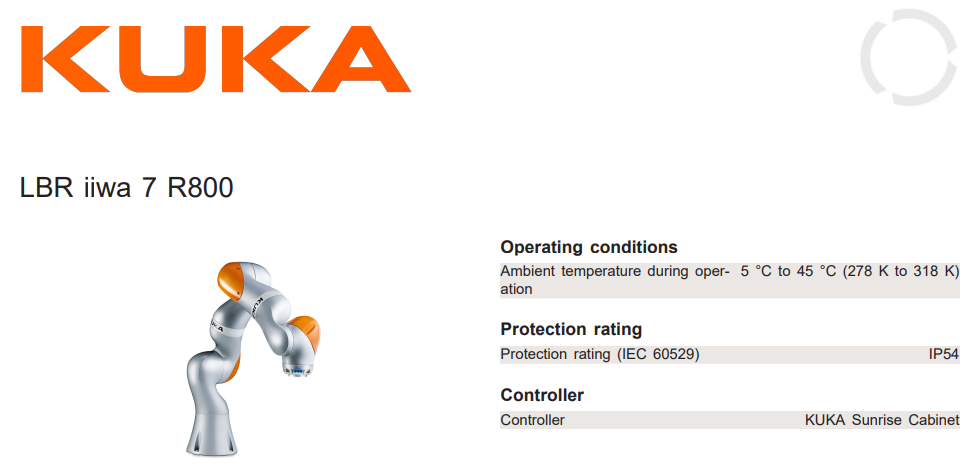}{kuka iiwa basic information}

\insertFig{0.9}{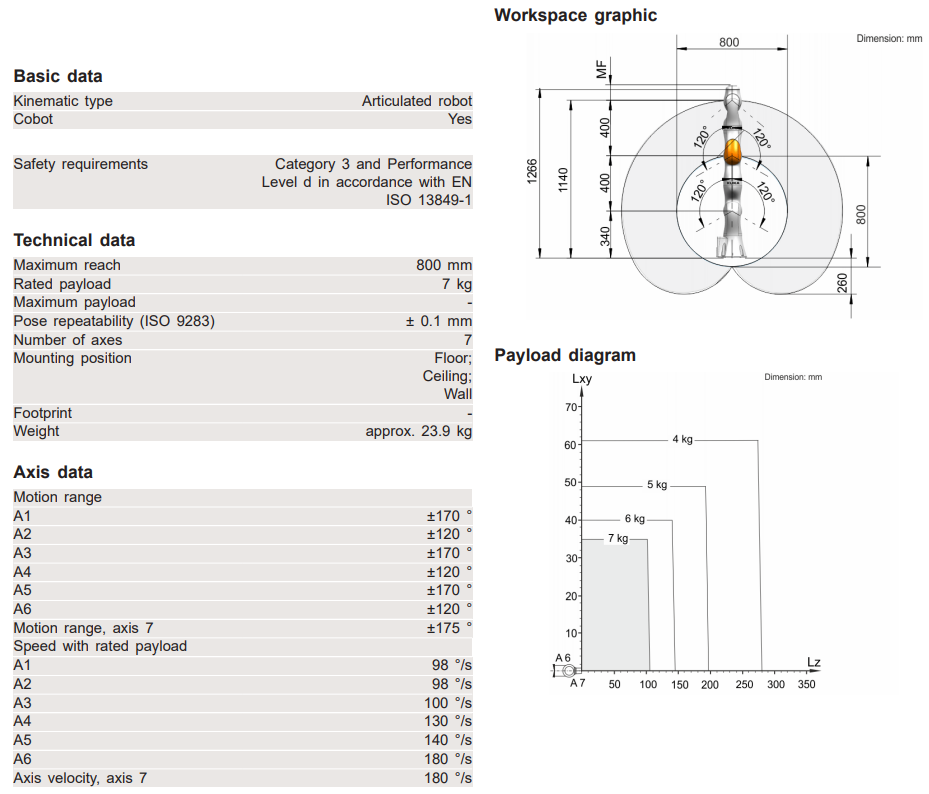}{kuka iiwa detail specifications}

\section{Episodic Differences Curves and Result}
\label{sec:appendix_ep}

\insertFig{0.9}{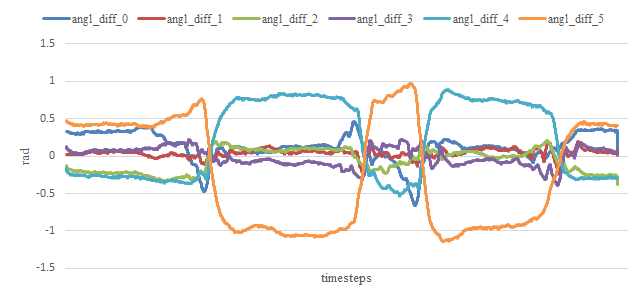}{Joint angle differences between the reference and robot motions ($video.LR$)}
\insertFig{0.9}{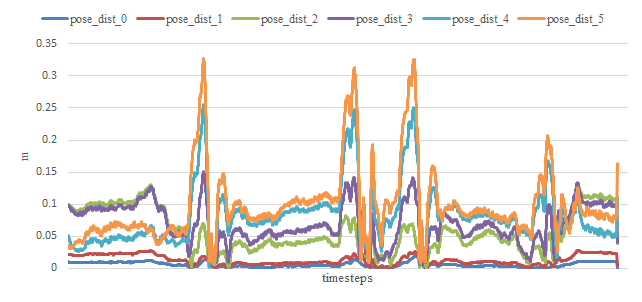}{Link position difference between the reference and robot motions ($video.LR$)}
\insertFig{0.9}{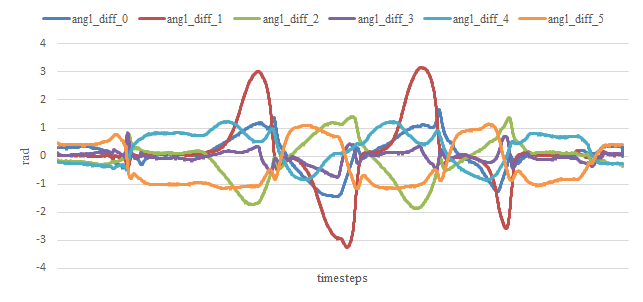}{Joint angle differences between the reference and robot motions ($video.UD$)}
\insertFig{0.9}{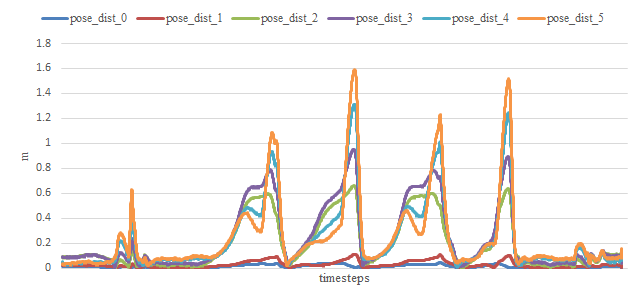}{Link position difference between the reference and robot motions ($video.UD$)}

\newpage

\end{document}